\documentclass[journal]{IEEEtran}

\usepackage[dvipsnames]{xcolor}

\usepackage{amsmath, amssymb}
\usepackage{times}
\usepackage{amssymb}

\usepackage{pifont}

\usepackage{siunitx}

\usepackage[utf8]{inputenc}

\usepackage{mathbbol}

\usepackage{esvect}

\usepackage[percent]{overpic}

\usepackage{tabu}

\usepackage{multirow}

\usepackage[overload]{empheq}

\usepackage{bm}

\definecolor{dg}{rgb}{0.1, 0.6, 0.2}       
\definecolor{b}{rgb}{0.0, 0.0, 1}          

\usepackage{epsfig}
\usepackage{float}
\usepackage{color}
\usepackage{booktabs} 
\usepackage{multirow} 
\usepackage{algorithm}      
\usepackage{algpseudocode}
\usepackage{amsfonts}
\usepackage{amsmath}
\usepackage{mathrsfs}

\usepackage{amsthm}
\usepackage{mathtools}

\usepackage{enumitem}       
\setenumerate[enumerate]{align=left}

\usepackage{graphicx}
\usepackage{subcaption}
\usepackage{caption}

\usepackage{balance}

\usepackage{multirow}

\usepackage[flushleft]{threeparttable}

\newcommand{\norm}[1]{\left\lVert#1\right\rVert}
\newcommand{\abs}[1]{\left\lvert#1\right\rvert}

\makeatletter
\newlength\tmp@\newlength\t@mp
\newcommand{\comp}[3]
  {\mathop{ \settowidth\tmp@{$\displaystyle\mathop{#1}^{#3}_{#2}$}
  \hbox to \tmp@{\hss \settowidth\t@mp{$\displaystyle #1$}\setlength\t@mp{.45\t@mp}
  $\displaystyle\mathop{#1}^{\hspace\t@mp #3}_{\hspace{-\t@mp}#2}$
  \hss} }}
\makeatother

\newcommand{\Int}[2]
{\int_{#1}^{#2}}

\DeclareMathOperator*{\argmin}{argmin}

\def\x{\mathrm{x}}

\def\G{\Gamma}

\def\R{\mathbb{R}}

\def\pos{\mathbf{p}}

\def\rot{\mathbf{R}}
\def\Tf{\mathbf{T}}
\def\trans{\mathbf{t}}

\def\V{\mathcal{V}}

\def\C{\mathcal{C}}

\def\Ob{\mathcal{O}}
\def\C{\mathcal{C}}
\def\E{\mathcal{E}}
\def\G{\mathcal{G}}

\def\bsP{\boldsymbol{P}}

\newtheorem{remark}{Remark}

\newcommand{\ul}[1]{\underline{#1}}

\usepackage{xstring}
\newcommand{\shorturl}[1]
{
    \href{https://#1}{\nolinkurl{#1}}
}

\usepackage{array}

\usepackage{tabularx}
\newcolumntype{Y}{>{\centering\arraybackslash}X}
\newcolumntype{R}{>{\raggedleft\arraybackslash}X}

\newcommand{\spd}{\mathrm{Sym}^{+}(3)}

\newtheorem{definition}{Definition}
\usepackage[pagebackref=false,breaklinks=true,colorlinks=true,citecolor=blue,linkcolor=blue]{hyperref}
\usepackage{cleveref}
\usepackage{cite}
\usepackage{amssymb}
\usepackage{mathtools, nccmath}

\def\objlabl{\ell}         
\def\objinst{\imath}     
\def\trace{\mathrm{tr}}
\title{\bf 
Probabilistic Scene Graphs:\\Hierarchical Representation and Real-time System
}
\author{Waqas Ali$^{1}$, Michele Antonazzi$^{1}$, Timon Homberger$^{1}$, Thien-Minh Nguyen$^{2}$, Lukas Rosenberger Schmid$^{3}$, Patric Jensfelt$^{1}$, Yixi Cai$^{1}$

\thanks{$^{1}$Authors are with the Division of Robotics, Perception, and Learning (RPL), KTH Royal Institute of Technology, Stockholm 114 28, Sweden.}

\thanks{$^{2}$Thien-Minh Nguyen is with the School of Mechanical and Mining Engineering (SMME), The University of Queensland, Brisbane, QLD 4072, Australia.}

\thanks{$^{3}$Lukas Rosenberger Schmid is with the Department of Computer Science and Artificial Intelligence, University of Technology Nuremberg, 90461 Nürnberg, Germany.}
}

\begin{document}

\maketitle


\begin{abstract}

3D scene graphs provide semantically rich and hierarchical representations for robot perception. However, existing systems do not maintain uncertainty as an explicit belief or propagate it through the operations that construct and refine the graph. We introduce \emph{Probabilistic Scene Graph} (PSG), a generalization of the conventional scene graph that represents a posterior over possible graphs, factorized into a discrete graph structure of entities, relations, and semantic attributes, and continuous states that ground them spatially, with uncertainty maintained over both components. Geometry is carried directly by the nodes rather than selected from a separately constructed metric map, so a metric map, where needed, follows from the graph rather than preceding it. We instantiate PSG's probabilistic spatial grounding with \emph{hierarchical graphs of Gaussians} (HGG): each object primitive is represented by a full-covariance Gaussian under a Normal-Inverse-Wishart belief, and the same parametrization applied recursively within a node yields a geometry graph that resolves its surface at finer resolution. We then build a mapping pipeline that preserves these beliefs throughout graph construction and refinement: a purely graph-based coarse-to-fine alignment registers observations by comparing node beliefs, while a nested Expectation-Maximization and factor-graph optimization jointly refines poses, object parameters, and internal geometry. Across six datasets spanning indoor RGB-D, outdoor LiDAR, and cross-modality deployment, HGG operates at sensor rate with near-constant memory and achieves state-of-the-art object accuracy and zero-shot graph alignment.

\end{abstract}

\section{Introduction}\label{sec:intro}
As robots are tasked with complex, instruction-driven objectives, a purely geometric map no longer suffices: acting on an instruction requires knowing not only where things are, but what they are and how they relate to one another. 3D scene graphs have emerged as a powerful representation for this, encoding semantic entities and their relations, often across multiple levels of abstraction~\cite{rotondi20263d}.

\begin{figure}
    \centering
    \includegraphics[width=\linewidth]{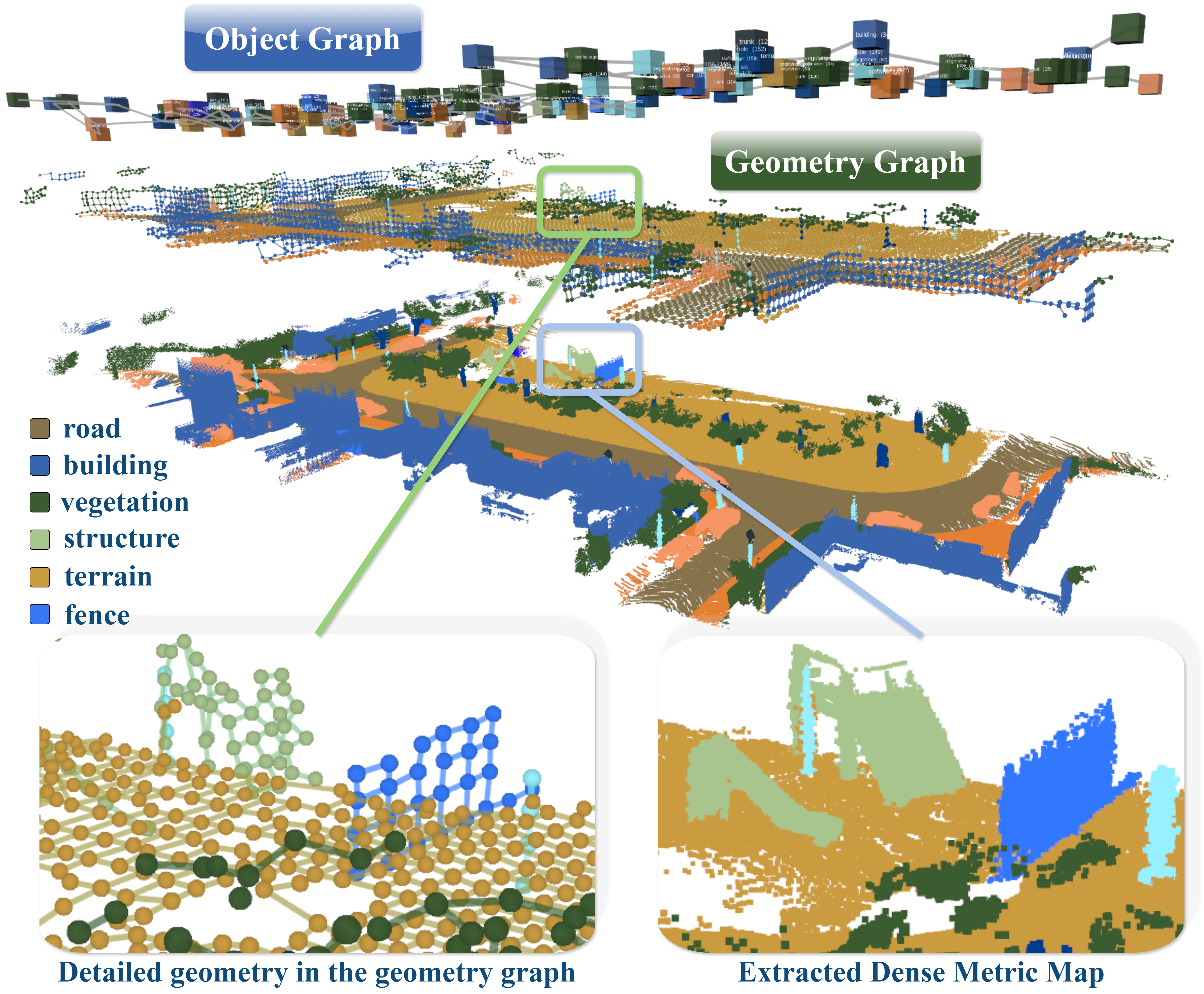}
    \caption{Our hierarchical probabilistic scene graph (HGG). \textbf{Top:} a symbolic object graph $G^o$, with a belief over centroid and full covariance, connected by relational edges. \textbf{Middle:} the geometry graph $G^g_i$ held inside each object, the same definition applied one level down, whose combined means and covariances provide a belief over continuous surface geometry. \textbf{Bottom:} detail of the geometry graph (left).  Realizations of dense semantic point clouds of the scene can be sampled from our PSG (right), comparable to the original measurements. Color denotes the semantic label.}

    \label{fig:scene_graph}
    \vspace{-0.6cm}
\end{figure}

Recent work has substantially advanced 3D scene graphs toward real-time, hierarchical, and open-vocabulary scene understanding. Kimera~\cite{rosinol2020kimera, rosinol2021kimera} integrated a dynamic scene graph into a visual-inertial SLAM pipeline, demonstrating semantic abstraction with full state estimation. Hydra~\cite{hughes2022hydra, hughes2024foundations} established the current state of the art, incrementally constructing a hierarchical scene graph, combining fast local updates with a dedicated optimization back-end. Open-vocabulary systems~\cite{gu2024conceptgraphs,werby23hovsg, maggio2024clio} showed that this hierarchy extends naturally to open-vocabulary semantics, replacing closed label sets with vision-language features without sacrificing structure. Collectively, these systems represent substantial progress.
However, uncertainty is not maintained as an explicit probabilistic belief in these scene graphs or propagated through the graph operations. Consequently, the uncertainty arising from noisy and partial observations is not consistently preserved throughout the mapping process.

This gap is at odds with the nature of the measurement process itself.
A robot does not have direct access to the true state of the scene; it receives noisy measurements drawn from a probabilistic generative process. Consider an object observed only partially and with noisy measurements. A conventional scene graph typically assigns it a point estimate of position and extent when the node is created. Once this estimate is stored, the graph no longer records how uncertain that estimate was. A later observation is therefore fused with a seemingly established object, even though the first observation may have been highly uncertain. This matters because uncertainty is not needed only at node creation. It affects every subsequent operation: deciding whether two observations correspond to the same object, registering a new graph to the map, and refining the object as more observations arrive. If uncertainty is discarded at the node, these operations cannot use it later.

We therefore propose the \emph{Probabilistic Scene Graph} (PSG), which represents a scene as a posterior over possible scene graphs rather than a single deterministic graph. Building on the scene-graph formulation of Rotondi et al.~\cite{rotondi20263d}, we decompose the graph into a discrete structure of entities, relations, and semantic attributes, and continuous states that ground these entities spatially. A PSG makes uncertainty an intrinsic part of the scene graph representation, maintaining beliefs over both components conditioned on the observations as the graph is constructed and refined. Within the continuous states, geometry is represented directly by each node: rather than grounding each entity through a separately constructed metric map, each node describes where its entity is and how far it extends. A metric map can therefore be derived from the scene graph when needed, rather than serving as a prerequisite for constructing it.

We instantiate this definition hierarchically around object primitives. At the object level, each node represents one object primitive, including both discrete objects such as chairs and desks and spatial structures such as walls or floor regions. These nodes form the object graph $\mathcal{G}^o$, while applying the same definition within each node yields a geometry graph $\mathcal{G}^g$ that describes the same object primitive at a finer resolution. Together, these two layers form the basis for the hierarchical representation of \Cref{fig:scene_graph}.

The probabilistic representation must be preserved through the process that constructs and refines the graph. Nodes are \emph{created} from partial observations, \emph{associated} with nodes already in the map to determine which observations correspond to the same object primitive, and \emph{corrected} as new observations arrive and the trajectory is revised. We therefore make each stage uncertainty-aware. Alignment compares node beliefs rather than point estimates, while a nested Expectation-Maximization and factor-graph optimization updates the beliefs as the map and trajectory are refined. In this way, the belief established at node creation is carried through association and correction rather than being discarded along the way. We summarize our contributions as follows:
\begin{itemize}
        \item \textit{The representation.} We introduce the probabilistic scene graph (PSG), a generalization of the conventional scene graph that represents a posterior over possible graphs rather than a single deterministic graph. The formulation makes uncertainty an intrinsic part of the representation, providing a principled basis for maintaining and updating uncertainty as the graph evolves.
        \item \textit{Creation.} We instantiate PSG with \emph{hierarchical graphs of Gaussians} (HGG): each object primitive is represented by a Gaussian under a Normal-Inverse-Wishart belief over its mean and covariance. Each object primitive further carries a geometry graph of finer Gaussians expressed in its object frame, whose aggregate statistics define the detailed object-level geometry.
        \item \textit{Alignment.} We extend the probabilistic representation to graph alignment with a purely graph-based coarse-to-fine method that registers newly observed graphs to the global graph without a dense metric map or learned embedding. Correspondences are determined from distributional geometry and structural context, then fused through evidence-weighted conjugate merging to preserve the probabilistic belief during map update.
        \item \textit{Correction.} We extend the probabilistic representation to graph correction with a nested factor-graph optimization and Expectation-Maximization that jointly refines poses and the parameters of object primitives while refining their internal geometry. A Fisher-information analysis shows that this separation incurs only a second-order loss in estimation efficiency relative to joint optimization.
\end{itemize}
We evaluate HGG on six datasets spanning indoor RGB-D, outdoor LiDAR, and cross-modality deployment, showing that the probabilistic representation yields state-of-the-art object accuracy and zero-shot graph alignment while operating in real time with near-constant memory. 
Language-grounded retrieval and navigation are built on the same probabilistic representation. Retrieval uses the spatial extent and relations encoded in the node beliefs to resolve relational queries, while navigation uses the uncertainty associated with the supporting nodes when planning through traversable space. In both cases, the robot reasons directly over the probabilistic scene graph rather than first reducing it to a deterministic map.

\vspace{-0.3cm}
\section{Related Work} \label{sec:related}
We organize related work along four lines that converge on the limitation that motivates this work. \Cref{sec:3d_scene} reviews 3D scene graphs, the representational paradigm this paper extends, tracing the field's progression toward real-time, open-vocabulary, and structurally integrated systems. \Cref{sec:ob-slam} reviews object-based SLAM, which approaches the same problem from an estimation-theoretic rather than a hierarchical-graph perspective and endows objects with first-class representational status. This line of work includes the parallel emergence of 3D Gaussian Splatting as a primitive. \Cref{sec:gr_align} reviews graph alignment, the problem of registering newly observed scene fragments into an existing map, which our mapping pipeline depends on directly. \Cref{sec:gr_opt} reviews graph-based optimization strategies for maintaining global consistency. 

Across all four lines of work, we identify the same structural limitation: node geometry is selected from a metric map that must be built first, and committed to a deterministic primitive that propagates no uncertainty. The field has reached a related diagnosis in a different register. Rotondi et al.~\cite{rotondi20263d} name uncertainty among the open challenges of dynamic scene graphs, observing that most approaches maintain a single estimate of the world state despite the uncertainty inherent in partial observability, ambiguous data association, and future scene evolution. 
Their open questions are about semantics and about time, while ours is about what both rest on. These pipelines decide that two observations are the same entity primarily from geometry, with semantics as a filter, and both are carried as point estimates. Neither geometry nor semantics has been treated as a belief that survives the operations performed on the node. We take up geometry, the channel every operation on a node passes through, and return to the semantic one in \Cref{sec:discussion}.

\vspace{-0.5cm}
\subsection{3D Scene Graphs} \label{sec:3d_scene}
3D scene graphs have emerged as the dominant paradigm for hierarchical, semantically rich environment representations in robotics. Armeni et al.~\cite{armeni20193d} proposed a multi-layered structure encoding unified semantics, 3D geometry, and camera relationships from RGB-D data.
Kim et al.~\cite{kim20193} introduced a sparse, object-centric formulation with objects as nodes and spatial-semantic attributes in the graph topology, an approach that proved effective for downstream task planning. Rosinol et al.~\cite{rosinol20203d} established the 3D Dynamic Scene Graph (DSG), a layered, actionable spatial perception model that integrates places, objects, and humans, which Kimera~\cite{rosinol2020kimera, rosinol2021kimera} grounded in a full visual-inertial SLAM pipeline. Wald et al.~\cite{wald2020learning} subsequently proposed learning the 3D scene graph structure directly from a point cloud and its instance segmentation, agnostic to the sensor or pipeline that produced the reconstruction.
Hydra~\cite{hughes2022hydra, hughes2024foundations} established the state of the art for real-time construction, incrementally building a layered scene graph spanning mesh, places, objects, and rooms, from visual-inertial data with fast local processing and slower global optimization. Despite these contributions, Hydra retains two structural limitations: it relies on dense metric-semantic meshes as its geometric substrate, and it enforces rigid, deterministic geometric constraints throughout, making it brittle to sensor noise and segmentation uncertainty.

The advent of large vision-language foundation models catalyzed a new generation of open-vocabulary scene graph systems. SceneGraphFusion~\cite{wu2021scenegraphfusion} demonstrated incremental scene graph prediction from RGB-D sequences using graph neural networks, with an attention mechanism to handle partial observations. ConceptGraphs~\cite{gu2024conceptgraphs} carried the open-vocabulary idea by fusing pre-trained CLIP and large language model features into 3D via multi-view association, producing object-centric representations that generalize to unseen semantic classes without task-specific fine-tuning. HOV-SG~\cite{werby23hovsg} extended this to a hierarchical open-vocabulary scene graph that covers object, room, and floor levels, each layer enriched with visual-language features. 
Clio~\cite{maggio2024clio} introduces the first real-time hierarchical open-set scene graph, using the open-set features to define task-relevant graph and object granularity.
FunGraph~\cite{rotondi2025fungraph} refined the granularity of objects to affordance-relevant object parts, enabling language-prompted interaction with functional elements rather than whole objects. LEXI-SG~\cite{kassab2026lexi} relaxed sensor requirements, demonstrating a dense monocular open-vocabulary scene graph mapping system by partitioning the scene into rooms and aligning per-room feed-forward reconstructions in a room-based factor graph. Collaborative dynamic scene graphs~\cite{greve2024collaborative} further extended the paradigm to autonomous driving, building shared semantic graphs among multiple vehicles for cooperative scene understanding.

A distinct research direction is the Situational Graphs (S-Graphs) family, which departs from the visual-inertial approaches by tightly coupling the scene graph into the SLAM factor graph itself. Bavle et al.~\cite{bavle2022situational} introduced S-Graphs as a jointly optimized three-layered graph combining pose keyframes, wall planes from 3D LiDAR, and topological room and corridor nodes. S-Graphs+~\cite{bavle2023s} extended this to an optimizable hierarchy spanning keyframes, walls, rooms, and floors, showing that architectural structure as first-class graph nodes improves trajectory accuracy and map consistency. The framework was later scaled to collaborative multi-robot mapping~\cite{fernandez2024multi}, extended with hierarchical semantic optimization in S-Graphs 2.0~\cite{bavle2025s}, and brought to the visual domain by vS-Graphs~\cite{tourani2025vs}. The family shows the value of structural semantics as optimization variables, but its nodes are deterministic primitives, including planar surfaces, room centroids, and floor levels, with no uncertainty model over their geometry.

Among LiDAR-based approaches, SGLC~\cite{wang2024sglc} employs a semantic graph of object nodes organized by class and spatial topology to drive coarse-to-fine loop closure detection. Wang et al.~\cite{wang2025leveraging} integrated a globally maintained semantic graph map constructed from foreground objects as an anchor structure running alongside a dense local point cloud. The graph map is used for place recognition and the dense local map for registration.

Across this line of work, the node primitive collapses the observation to a point estimate at creation, and no system propagates distributional uncertainty over geometry through association, registration, or optimization. We instead make each node a Gaussian with full covariance over the object's extent, maintaining a belief over that Gaussian which is carried through association and correction.
\vspace{-0.2cm}
\subsection{Object-based SLAM}\label{sec:ob-slam}

Object-based SLAM elevates semantic object instances to first-class landmarks, enabling richer scene understanding and more discriminative loop closure than feature-point or volumetric backends alone. SLAM++~\cite{salas2013slam++} established the paradigm using pre-built CAD models, later relaxed by CubeSLAM~\cite{yang2019cubeslam} and QuadricSLAM~\cite{nicholson2018quadricslam}, which estimate object pose and extent as cuboids and dual quadrics directly from detections, removing the closed-vocabulary constraint. 
Bowman et al.~\cite{bowman2017probabilistic} maintain a distribution over which landmark generated each measurement, preserving uncertainty over object identity. The landmark itself remains a point, however: once association is resolved, no distributional state remains over the object's extent or shape.
NodeSLAM~\cite{sucar2020nodeslam} instead reconstructs per-instance shape using neural priors. OA-SLAM~\cite{zins2022oa} and VOOM~\cite{wang2024voom} integrate ellipsoidal and dual-quadric object landmarks into conventional bundle adjustment, the latter using coarse object correspondence to guide fine point-feature association. ObVi-SLAM~\cite{adkins2024obvi} extends this paradigm to long-term, multi-session operation, and Wu et al.~\cite{wu2023object} provide a systematic treatment of how object-level maps support downstream manipulation and navigation tasks.

A parallel line of work replaces parametric object shapes with 3D Gaussian Splatting (3DGS)~\cite{kerbl20233d}, whose anisotropic Gaussians (mean, full covariance, opacity, view-dependent color) enable real-time, directly-manipulable radiance fields. GS-SLAM~\cite{yan2024gs}, Gaussian Splatting SLAM~\cite{matsuki2024gaussian}, SplaTAM~\cite{keetha2024splatam}, and Photo-SLAM~\cite{huang2024photo} adapted this representation to real-time dense tracking and mapping, while Gaussian-LIC~\cite{lang2025gaussian} and MM3DGS SLAM~\cite{sun2024mm3dgs} extended it to LiDAR-camera fusion, and SGS-SLAM~\cite{li2024sgs}, NEDS-SLAM~\cite{ji2024neds}, and SemGauss-SLAM~\cite{zhu2025semgauss} augmented individual Gaussians with semantic labels or learned features. 
Across this literature, however, the Gaussian remains a sub-object primitive: thousands are aggregated into a dense representation, and none is used to represent a complete semantic object in its own right~\cite{zhu20243d, tosi2026nerfs, xuan2025survey}.

The two lines converge on the same limitation from opposite directions. Object-based SLAM fixes geometry as a parametric shape that absorbs no noise, and where that shape is already an ellipsoid~\cite{nicholson2018quadricslam, zins2022oa}, it is estimated as a point rather than maintained as a belief. 
In 3DGS-SLAM the Gaussian is a rendering primitive: its covariance is a splat footprint fitted to photometric error rather than a statement about an object's extent, and no belief is maintained over it.
Our method closes the gap by representing each object's position and extent with a full-covariance Gaussian and recovering finer shape by applying the same primitive within the node rather than over a dense representation, so its geometry graph Gaussians are attributes of a single object rather than an unsegmented reconstruction of the scene.

\subsection{Graph Alignment}\label{sec:gr_align}

Graph alignment seeks a consistent node-level correspondence between two graphs with partial overlap, differing topology, or heterogeneous attributes; in 3D scene mapping, it underpins place recognition and the registration of new scene fragments into a global representation. Classical formulations cast matching as quadratic assignment solved via spectral relaxation~\cite{leordeanu2005spectral, cho2010reweighted}, but are computationally expensive and ill-suited to the partial-overlap regime ubiquitous in robotics. The Weisfeiler-Lehman kernel~\cite{shervashidze2011weisfeiler} yields graph-level similarity rather than node correspondence, though its iteration can be adapted into a per-node structural descriptor, as we do in \Cref{sec:fine_corr}.
For 3D scene graphs, SGAligner~\cite{sarkar2023sgaligner} learns a joint multi-modal contrastive embedding for node matching, followed by a decoupled point cloud registration step. SG-PGM~\cite{xie2024sg} fuses scene graph and point cloud geometry in a shared encoder with differentiable top-k partial matching to handle partial overlap explicitly. ROMAN~\cite{peterson2025roman} aligns open-set object maps by solving a densest edge-weighted clique over a pairwise consistency graph, without an initial pose estimate. SG-REG~\cite{liu2025sg} fuses open-set semantic, topological, and shape cues into a single compact node feature, targeting the generalization gap of learning-based registration. OpenSGA~\cite{chen2026opensga} demonstrated open-world 3D scene graph alignment, extending the domain from indoor closed-set scenes to arbitrary environments with novel semantic classes.

All of these methods, spanning contrastive embeddings~\cite{sarkar2023sgaligner}, fused geometric-semantic encoders~\cite{xie2024sg, liu2025sg}, and open-world matching~\cite{chen2026opensga}, resolve the correspondence over deterministic node attributes. We instead make the correspondence cost itself a distributional distance between node beliefs, evaluated under a structure valid on the manifold that their covariances occupy, and requires neither a point cloud encoder nor an embedding trained for graph alignment.

\subsection{Graph Correction}\label{sec:gr_opt}
Graph correction is crucial to achieving locally and globally consistent maps.
Current strategies perform the correction by reintegrating volumes~\cite{dai2017bundlefusion}, realigning submaps~\cite{reijgwart2019voxgraph}, optimizing unstructured surfels~\cite{whelan2016elasticfusion}, or applying deformation graphs to meshes~\cite{rosinol2021kimera}. 
However, these methods focus on optimizing dense representations, so object-level entities are corrected only as a consequence of the surface or volume they were derived from rather than being estimation variables in their own right. 
Hydra~\cite{hughes2022hydra} introduces a graph-based backend that optimizes the object and place layers alongside the meshes, yet remains limited to rigid geometric constraints, which hinders its robustness to sensor noise and perception uncertainty.
Factor graph representations have also been proposed for scene-level uncertainty reasoning beyond pure pose estimation. 
Khronos~\cite{Schmid-RSS24-Khronos} extends Hydra to object-level reconstruction and uses a factor graph to disambiguate object association from scene changes.
Millan-Romera et al.~\cite{millan2024generation} demonstrated that factor graphs can represent distributional uncertainty over semantic concepts in factorized 3D scene graphs, motivating the use of probabilistic factor formulations at the scene graph level. Their formulation, however, covers semantic concepts rather than object geometry, and operates offline rather than incrementally.
Every strategy above either optimizes a dense map or treats object nodes as points in Euclidean parameter space. Our nested framework optimizes poses and object parameters jointly in a factor graph and refines each object's internal geometry conditioned on them, with residuals formulated on the manifold of symmetric positive-definite matrices ($\spd$) rather than its Euclidean embedding.


\section{Foundation of a Probabilistic Scene Graph}\label{sec:foundation}
\subsection{Distribution over Scene Graphs}
Across the systems of \Cref{sec:related}, a scene graph is typically the output of a two-stage pipeline
\begin{equation}
  \mathcal{Z} \;\xrightarrow{\;\text{reconstruct}\;}\; \mathcal{M}
  \;\xrightarrow{\;\text{abstract}\;}\; \mathcal{G},
  \label{eq:pipeline}
\end{equation}
carrying sensor measurements $\mathcal{Z}$ to a metric map $\mathcal{M} \subseteq \mathbb{R}^3$ and then to the tuple $\mathcal{G} = (V, V_g, V_f, E, E_f, [L])$ -- a node set $V$, a grounding of nodes on the map $V_g$, node attributes $V_f$, an edge set $E$, edge attributes $E_f$, and an optional assignment of nodes to layers $[L]$ -- consolidated by Rotondi et al.~\cite{rotondi20263d}. Each node $v_i \in V$ represents an entity grounded in 3D space (an object, an object part, or a region), and the grounding $V_g: V \to \{0,1\}^{\mathcal{M}}$ assigns it the subset of map points that the entity occupies. 
This pipeline design is appealing, as it deals effectively with two of the primary challenges in 3D scene graph estimation.
First, scene graphs are inherently discrete, symbolic structures, whereas the scene geometry is a continuous entity.
Thus, $\mathcal{M}$ creates a continuous reconstruction of geometry, from which the symbolic graph can be abstracted.
Second, if a map $\mathcal{M}$ is maintained, especially in the presence of odometry drift and other perceptual challenges, associations $V_g$ of $V$ into $\mathcal{M}$ provide an effective solution to the hard problem of spatially grounding the scene graph.

However, this separation has notable limitations, as measurements are collapsed to a single belief and a one-way dependence of the scene graph on the geometric map is created.
We address these limitations of \eqref{eq:pipeline} as follows.
First, we estimate a posterior over possible scene graphs rather than a single scene graph. 
Let $\mathbb{G}$ denote the set of scene graphs, with each graph decomposed into a discrete structure $S=(V,E,V_f)$, for example describing its entities, relations, and semantic attributes, and continuous states $\theta$.
While any attributes of the graph may be continuous, without loss of generality, we here focus on capturing the spatial grounding in $\theta$, an essential component of any probabilistic scene graph.
The resulting posterior $P(S,\theta|\mathcal{Z})$ thus captures uncertainty in both components, symbolic and spatial, allowing it to be maintained as the graph is created, aligned, and corrected: each operation takes a posterior to a posterior rather than reading out a point estimate.
Second, within the continuous states $\theta$, we change how geometry is captured. In the classical formulation, $V_g$ grounds each entity by selecting the subset of points in the metric map $\mathcal{M}$ that it occupies. We instead propose a scene-graph-first approach, where each node carries its geometric state directly: $\theta_i$ describes where the entity is, how far it extends, and how its surface is distributed. 
This removes a prerequisite, not a modality. 
The map $\mathcal{M}$ may be a point cloud, a mesh, or a splat model, but $V_g$ can only ground an entity in a map that has already been built and is thereafter kept.
In our formulation $\theta$ is itself a distribution over scene geometry, so a metric map follows from the graph rather than preceding it and can be be sampled from $\theta$ if desired. This leads to the following formal definition of a probabilistic scene graph.

\begin{definition}[Probabilistic scene graph]
\label{def:psg}
A probabilistic scene graph  is a posterior $P(S, \theta | \mathcal{Z})$ over $\mathbb{G}$,
\begin{equation}
  \mathcal{Z} \;\longrightarrow\;
  P(S, \theta | \mathcal{Z})
  = \underbrace{\pi(S | \mathcal{Z})}_{\text{symbolic}}
    \underbrace{B(\theta | S, \mathcal{Z})}_{\text{geometry}}
  \;\in\; \mathcal{P}(\mathbb{G}).
  \label{eq:factorization}
\end{equation}
\end{definition}
\noindent Here, $\pi(S|\mathcal Z)$ represents the posterior over the discrete structure, while $B(\theta|S,\mathcal Z)$ represents the conditional belief over the continuous states. Written field by field against the classical tuple, and suppressing the common conditioning on $\mathcal{Z}$, Definition~\ref{def:psg} becomes $\mathcal{G}_P$,

\small
\begin{equation}
  \mathcal{G}_P = \bigl(\, \pi(V), B(\theta|S), \pi(V_f|V, E), \pi(E|V), \pi(E_f|V,E), \pi(L) \,\bigr)
  \label{eq:psg}
\end{equation}
\normalsize
with symbolic entries $\pi(S) = \pi(V)\,\pi(E|V)\,\pi(V_f|V, E)$, and $\pi(E_f|V,E)$, and continuous belief $B(\theta|S)$. 

The probabilistic entries in $\mathcal{G}_P$ replace the corresponding deterministic fields in $\mathcal{G}$ with distributions over the values they could take: $\pi(V)$ over which entities exist, $\pi(E|V)$ over which are related, $\pi(V_f|V, E)$ over what they are, $\pi(E_f|V, E)$ over how they are related, $B(\theta|S)$ over their geometry, and $\pi(L)$ over their organization in hierarchical layers. 
The definition leaves open the parametric family, the granularity of the entities, and which groundings are realized as beliefs rather than point estimates.

\begin{figure*}
    \centering
    \includegraphics[width=\linewidth]{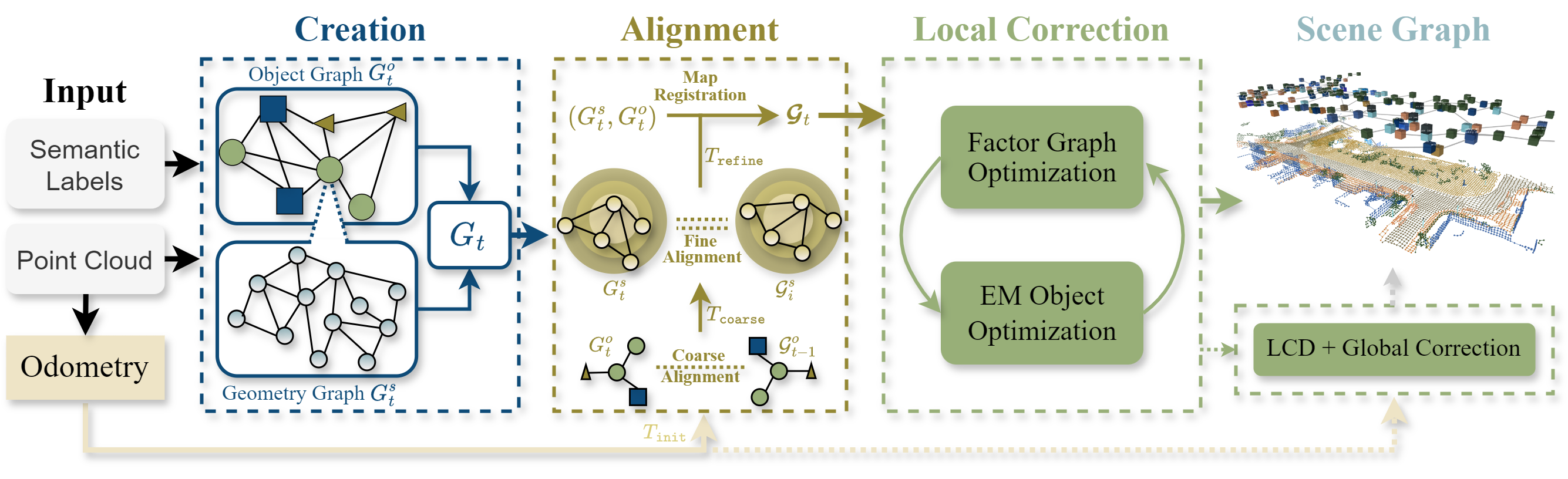}
    \vspace{-0.75cm}
    \caption{System overview. Given a semantic point cloud frame and odometry, we create a hierarchical local graph $G_t$, align it with the global graph $\G_{t-1}$ through coarse-to-fine graph alignment, and update the global graph $\G_{t}$ through map registration. The resulting representation is refined through nested optimization within a sliding window, followed by global optimization.}
    \label{fig:system}
    \vspace{-0.4cm}
\end{figure*}

\begin{table}[tb]
\centering
\caption{Notation for the Probabilistic Scene Graph}
\label{tab:not}
\setlength{\tabcolsep}{8pt}
\begin{tabular}{ll}
\toprule
\textbf{Notations} & \textbf{Description}  \\ \midrule
$t$ & Frame index \\
$G_t$ & Local scene graph created from frame $t$   \\
$\G_t$ & Global scene graph registered up to frame $t$ \\
$G^o$, $\G^o$ & Superscript $o$ denotes the object graph\\
$G^g_i$, $\G^g_i$ & Superscript $g$ denotes the geometry graph belonging to \\
&$i^{\text{th}}$-node of the object graph $G^o$, $\G^o$\\ 
\bottomrule
\end{tabular}
\vspace{-0.6cm}
\end{table}

\vspace{-0.25cm}
\subsection{From Definition to Parametrization}\label{sec:higogo}
The PSG formulation in \Cref{def:psg} is general and does not prescribe how uncertainty is represented or how the scene graph is organized. 
However, as a first step, any PSG requires probabilistic spatial grounding that generalizes 
across levels of granularity, from a whole entity to the surface within it,
and that can be efficiently manipulated and updated while maintaining its probabilistic properties.
We thus propose \emph{hierarchical graphs of Gaussians} (HGG) as a representation for probabilistic spatial grounding.

In HGG, the same parametrization is applied in each layer.
We parameterize $\theta_i$ for each symbolic node $i$ as a density over space, described by a mean position $\mu_i$ and a covariance $\Sigma_i$, which specify where the node is and how far it extends.
Among all densities on $\mathbb{R}^3$ with a given mean and covariance, the Gaussian has maximum entropy~\cite{cover1991elements}, making $\mathcal{N}(\mu_i,\Sigma_i)$ the least committed choice consistent with these two quantities. 
We use a full covariance rather than an isotropic or axis-aligned one to capture directional variation in the extent of an entity.
Consequently, $\Sigma_i$ lies on the manifold $\spd$ rather than in a flat parameter space. 
We therefore perform comparison, interpolation, and refinement using operations that respect this manifold structure.

Note that the pair $(\mu_i,\Sigma_i)$ describes the physical node extent and is not itself an uncertainty. 
Instead, the PSG formulation requires the node to hold a belief over these parameters, represented here as a distribution $P(\mu_i,\Sigma_i|\mathcal{Z})$ on $\mathbb{R}^3\times\spd$.
We maintain this in conjugate form as a Normal-Inverse-Wishart posterior~\cite{murphy2007conjugate, gelman1995bayesian}.
This choice also provides two useful properties: the dispersion of the belief is governed by the number of measurements attributed to the node, and the posterior remains proper at every finite number of observations. 
\Cref{sec:Initialization} introduces these parameters and their update in more detail.

Since spatial relations are an essential component of most scene graphs, we further introduce a set of probabilistic spatial edges $E(V)$.
In this formulation, $E$ denotes the neighborhood between nodes within a layer $L$.
Their corresponding edge attributes $E_f$ are weights induced from the node geometry and therefore inherit uncertainty from the beliefs of the nodes they connect. 

Finally, a mean and covariance describe where, for example, an object is and how far it extends, but not its detailed shape. 
We therefore introduce a geometry graph $\mathcal{G}^g_i$ capturing the continuous surface of node $i$, which is simply another layer of the HGG representation introduced above.
The geometry graph provides this finer representation: its nodes cover parts of the same entity, and their combined mean and covariance define the probabilistic distribution over surface geometry. Thus, the symbolic node and its geometry graph represent the same object at two resolutions rather than as two separate representations. The finer graph also carries spatial relations between its nodes, providing the structure used for fine alignment and for propagating geometric corrections between neighboring parts.

\vspace{-0.25cm}
\subsection{Probabilistic Scene Graph Instantiation}

In this work, we demonstrate PSGs in one instantiation of \Cref{def:psg}, although many others are possible.
We build a hierarchical PSG spanning a symbolic and a continuous layer; an object graph $\mathcal{G}^o$ whose nodes are the entities of the scene, and within each of those nodes a geometry graph $\G^g$ describing the same entity at a finer resolution.
\Cref{tab:not} describes the notations.

We model the geometric factor $B(\theta|S,\mathcal{Z})$ probabilistically, while the symbolic factor $\pi(S|\mathcal{Z})$ is instantiated deterministically. 
This is a design choice reflecting our focus on spatial grounding rather than a restriction of the PSG formulation; the broader formulation also allows uncertainty in the symbolic components, as discussed in \Cref{sec:discussion}.
Similarly, each layer contains a set of spatial edges $E$ and features $E_f$ denoting probabilistic neighborhood relations. 
Again, additional semantic edges or features can readily be added to the PSG formulation.

\vspace{-0.35cm}
\section{Methodology}\label{sec:system}
Our approach to estimate the HGG introduced above in real-time is summarized in \Cref{fig:system}.
The system takes as input a semantic point cloud and odometry and processes them through three stages: creation, alignment, and correction. 
Creation (\Cref{sec:Initialization}) constructs a local hierarchical graph $G_t$ from each input frame, expressed in the local body frame. The graph is thus the unit of measurement rather than a product of accumulation. Creating it per frame bounds the front-end cost by one frame rather than by the mapped volume, letting $\kappa^o$ record what a single observation actually saw, and makes every downstream stage graph-to-graph and hence indifferent to the sensor.
Alignment (\Cref{sec:Alignment}) establishes coarse-to-fine correspondences between the local graph $G_t$ and the global graph $\G_{t-1}$, using the odometry $\Tf_{\mathrm{init}}$ to initialize the registration and refining it to obtain $\Tf_{\mathrm{refine}}$. Map update then transforms $G_t$ by $\Tf_{\mathrm{refine}}$ and merges it into $\mathcal{G}_{t-1}$ to produce the updated global graph $\mathcal{G}_t$.  Correction (\Cref{sec:Optimization}) refines the poses and the representation at two levels; locally within a sliding window using nested optimization, followed by global loop closure detection and pose-graph optimization to enforce global consistency.


\subsection{Creation}\label{sec:Initialization}
We create a local scene graph, including an object graph and the geometry graphs within its nodes, from a point cloud and its semantic labels. In this process, the scene graph components in \eqref{eq:psg} are instantiated with only the geometric factor $B$ represented as a belief: $\pi(V)$ and $\pi(V_f | V, E)$ are taken from instance segmentation as point estimates,  $\pi(E | V)$ from a deterministic neighborhood rule and the edge attributes $E_f$ are computed from the beliefs of the two nodes an edge connects. 
We group the semantic point cloud into object instances $\Ob_i$ using Euclidean clustering~\cite{rusu2010semantic, rusu20113d}, and use $\bsP_i$ to represent the set of points of each object $\Ob_i$.

\subsubsection{Object Graph}
The objects $\Ob_i$ form the node set $\V^o$ of the object graph $G^o$. 
Each node holds one joint distribution over an object's position and extent and the robot's confidence in it. The parameters $(\mu_i^o, \Sigma_i^o)$ describe the physical object: $\mu_i^o$ is its position and $\Sigma_i^o$ its spatial extent. Together with $\kappa_i^o$ they parameterize a single distribution over that object, the geometric factor $B$ of \eqref{eq:psg} in the conjugate form,
\begin{equation}\label{eq:niw}
    P(\mu_i, \Sigma_i | \mathcal{Z}_i) = \mathrm{NIW}\left(\mu, \Sigma | \mu_i^o, \kappa_i^o, \kappa_i^o\Sigma_i^o, \nu_i\right),
\end{equation}
where $\mathcal{Z}_i$ denotes the observations attributed to $\Ob_i$, $\kappa_i^o \in \mathbb{R}^+$ is the pseudo-count, and $\nu_i = \kappa_i^o + n_0$ with $n_0 = d+1 = 4$ for $d = 3$ dimensions.

At creation, the parameters are read off the sufficient statistics of $\bsP_i$: $\mu_i^o$ and $\Sigma_i^o$ are its sample mean and covariance, and $\kappa_i^o = |\bsP_i|$, so that each measurement counts as one unit of evidence. Two properties follow. The offset $n_0 = d+1$ makes the posterior expectations exact, $\mathbb{E}[\mu|\mathcal{Z}_i] = \mu_i^o$ and $\mathbb{E}[\Sigma|\mathcal{Z}_i] = \Sigma_i^o$, so the belief is centered on the measured position and extent rather than on a contracted version of them, and the single Gaussian $\mathcal{N}(\mu_i^o, \Sigma_i^o)$ that alignment and correction act on is formed from these two expectations directly. The width of the belief is then governed by $\kappa_i^o$ alone: the position is localized to within $\Sigma_i^o/\kappa_i^o$, without the extent being presumed known, and the extent is resolved at the same $1/\kappa_i^o$ rate. This is what makes $\kappa_i^o$ a measure of how much of the object the sensor has seen. An object glimpsed once at range thus remains uncertain in both position and extent until further views accumulate. The complete definition of the object is:
\begin{equation}
    \Ob_i = (\bsP_i, \objlabl_i, \objinst_i, G_i^g, (\mu_i^o, \Sigma_i^o), \kappa_i^o).
\end{equation}
Here $G_i^g$ is the geometry graph of \Cref{sec:higogo}, built in \Cref{sec:subsym}. Note that the point set $\bsP_i$ is held only while $\Ob_i$ lies within the sliding window of \Cref{sec:Optimization} and is released once the window passes, so the memory footprint of a node in the persistent map is its parameters alone.

We define edges based on relative neighborhood graph topology (RNG)~\cite{toussaint1980relative}, which yields a sparse connected topology and is widely used in pattern recognition~\cite{wang2021comprehensive, gyllensten2015navigating}. The edge set $\E^o$ contains $(i,j)$ when no third object $\Ob_k$ is closer, in Euclidean distance between centroids, to either $\Ob_i$ or $\Ob_j$ than they are to each other. 
However, since spatial edge attributes $E_f$ should capture the similarity of two nodes, a weight $w_{i,j}$ computed from the centroids alone would defy the probabilistic nature of our approach.
We instead define an edge weight that makes full use of $\theta$.

To measure geometric similarity, we use $S_1$ as the unified distance of~\cite{abou2010designing}, which sums a term in the centroids~\cite{bhattacharyya1943measure} and a term in the covariances, the Affine Invariant Riemannian Metric~\cite{pennec2006riemannian}:

\small
\begin{align}\label{eq:metric}
    S_1 = \norm{\mu_i^o - \mu_j^o}_{\mathbf{S}^{-1}}
    + \sqrt{\sum_{k=1}^{d} \left(\ln \lambda_k\right)^2},
\end{align}
\normalsize
where $\mathbf{S} = \frac{1}{2}(\Sigma_i^o + \Sigma_j^o)$ and $\lambda_1, \dots, \lambda_d$ denotes the generalized eigenvalues $\lambda_k$ solving $\Sigma_i^o \x_k = \lambda_k \Sigma_j^o \x_k$. The first term of \eqref{eq:metric} is the separation of the centroids scaled by their combined extent, so that a given separation counts for less between large objects than between small ones. The second is the geodesic distance between $\Sigma_i^o$ and $\Sigma_j^o$ on $\spd$. An edge weight therefore reads the entities as distributions rather than as points: it grows with separation measured against the objects' own extent, and with disagreement between their shapes.

In addition, we define a semantic term $S_2$ based on a label consistency check:
\begin{equation}\label{eq:tpm}
  S_2 =
    \begin{cases}
      0 & \text{if }  \objlabl_i = \objlabl_j  \\
      1 & \text{otherwise}
    \end{cases}.
\end{equation}
The final edge weight $w_{i,j}$ combines the means, the covariances, and the semantic labels:
\begin{equation}\label{eq:rel_weights}
    w_{i,j} = c\frac{S_1}{S_{\max,1}} + (1-c)S_2,
\end{equation}
where $S_{\max,1}$ is the maximum of $S_1$ over $\E^o$, used for normalization.
The weights are collected in $\mathcal{W} = [w_{i,j}]_{(i,j) \in \E^o}$, the edge attribute of the object graph $G^o_t \triangleq (\V^o, \E^o)$. 
Intuitively,  $w_{i,j}$ grows with separation, shape disagreement, and label mismatch, thus marking a more dissimilar pair.  

\subsubsection{Geometry Graph}\label{sec:subsym}
Within the $i$-th node of the object graph $G_t^o$, we recursively partition the point cloud $\bsP_i$ using a K-D tree into $N_i$ subsets $\{\bsP_{i,k}\}_{k=1}^{N_i}$, with the recursion terminating when each leaf contains at most a preset number of points. These subsets instantiate the nodes of the geometry graph $G_i^g$. For each subset $\bsP_{i,k}$, we estimate its parameters $(\mu_k^g,\Sigma_k^g)$ as the mean and covariance, expressed in the \emph{object frame}. The object frame is centered at the object centroid $\mu_i^o$ and oriented such that its axes are aligned with the eigenvectors of the object covariance $\Sigma_i^o$. The pseudo-count is defined as $\kappa_k^g = |\bsP_{i,k}|$, corresponding to \eqref{eq:niw} applied to the subset $\bsP_{i,k}$. It provides the initial evidence weight of each geometry node and is subsequently re-estimated during EM optimization (\Cref{sec:em}). Applying the RNG condition to the geometry nodes yields the edge set $\E_i^g$. Expressing the geometry graph in the object frame isolates its internal geometry from the object's global position and orientation, providing the local coordinate system used for subsequent geometric refinement.

The construction is identical for all nodes, including things and stuff classes. 
Note that the resolution is governed by the granularity parameter and the number of observations in the measurement, which thus automatically prioritizes detailed close-up observations. At the same time, the background is described at the same resolution.
This is possible because HGG is efficiently and sparsely parametrized.

\subsubsection{Recursive Hierarchy}
More generally, the same construction can be applied at a coarser level, since the PSG formulation does not prescribe the granularity at which nodes are represented. Groups of object primitives that belong together, such as a room indoors or a car park outdoors, can therefore be represented as a higher-level node. Its mean, covariance, and belief are obtained by merging its constituent nodes with \Cref{eq:merge}, while its edges are constructed from the resulting parameters using \Cref{eq:rel_weights}. Further levels introduce the additional question of which object primitives belong together, i.e., uncertainty in the symbolic structure, which is left deterministic in our current instantiation and discussed in \Cref{sec:discussion}.


\subsection{Alignment}\label{sec:Alignment}
Alignment registers the graph $G_t$ of the current frame into the global graph $\G_{t-1}$ and fuses it with what is already there. It is the first stage that compares beliefs rather than constructing them. Its cost functions compare distributions directly. We propose a coarse-to-fine method that follows the hierarchy (\Cref{fig:system}). The method applies to any number of levels, with each level gating the search at the one below; in our instantiation two suffice, since the object graph is sparse, so correspondences can be efficiently sought globally over every semantically valid pair, giving $\Tf_\text{coarse}$. The geometry graphs are far denser, but need only be searched within the pairs the object graph has already matched, giving the refined $\hat{\Tf}_\text{fine}$. The matched nodes are then merged into the map, and the two stages together make the estimate robust to odometry error and to geometric inconsistency from sensor noise or partial observation.

\begin{figure*}
    \centering
    \includegraphics[width=0.875\linewidth]{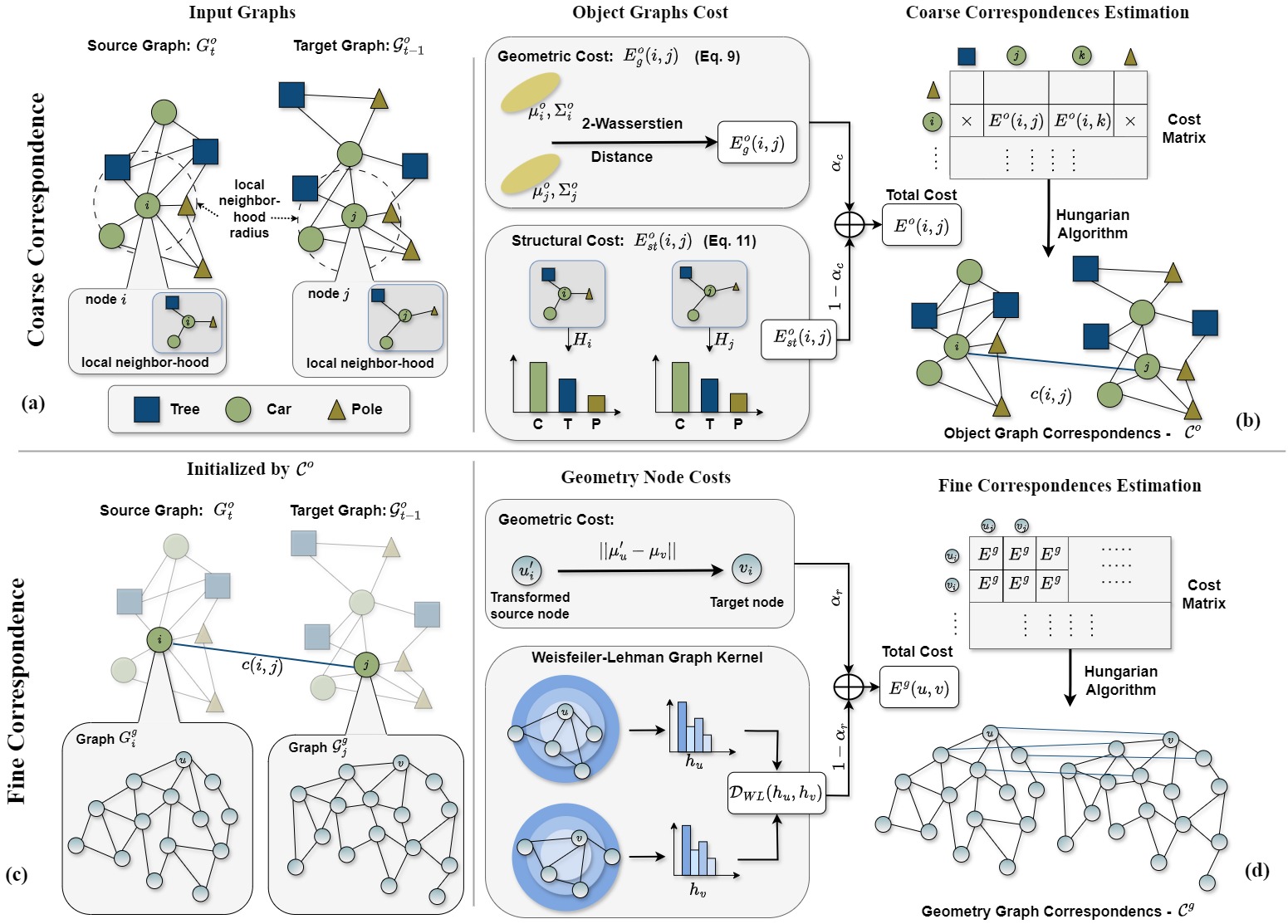}
    \vspace{-0.15cm}
    \caption{Coarse-to-fine graph alignment between the local graph $G_t^o$ and the global graph $\mathcal{G}_{t-1}^o$. 
\textbf{Top:} object-node-level correspondences combine geometric and structural costs, solved by the Hungarian algorithm. 
\textbf{Bottom:} the resulting correspondences initialize geometry-level matching, which combines spatial and Weisfeiler--Lehman structural costs.}
    \label{fig:corres}
    \vspace{-0.65cm}
\end{figure*}

\subsubsection{Coarse Correspondence Estimation}\label{sec:cors_corr}

To initialize the alignment pipeline between graphs $G_{t}^o$ and $\G_{t-1}^o$, we seek to build an optimal set of node-to-node correspondences $\C^o = \{ (i,j) | i\in \V_{t}^o, j \in \V_{t-1}^o \}$, each pair $(i,j)$ matching a node of $G_{t}^o$ to a node of the map $\G_{t-1}^o$. The coarse correspondence estimation $\mathcal{C}^o$ is illustrated in \Cref{fig:corres}(a)-(b). 

The main problem is formulated as a global minimization problem with hard semantic constraints, i.e., nodes that are not of the same label will not make a correspondence. For each valid pair of nodes, we calculate a cost with a combination of geometric attributes and topological context, as shown in \Cref{fig:corres}(b). To estimate the geometric cost $E_g^o(i, j)$, where $i \in \V_{t}^o$ and $j \in \V_{t-1}^o$, we quantify the position and extent discrepancy by computing the 2-Wasserstein distance~\cite{villani2008optimal}. 
Note that neither of the standard alternatives leverages this information: a cost on $\mu^o$ decides the association purely on the centroid and IoU between fitted volumes vanishes as soon as two objects stop overlapping, both of which are highly sensitive to pose errors and sensing noise. 
The 2-Wasserstein distance instead uses $\mu^o$ and $\Sigma^o$ together, and between Gaussians it can be computed in closed form, efficiently enough to be evaluated at the association rate:

\small
\begin{align} \label{eq:wass}
     E_g^o(i, j) &= \sqrt{\norm{\mu_i^o - \mu_j^o}^2 + D_{i,j}},
     \\\nonumber
          D_{i,j} &\triangleq
     \trace
     \left[
     \Sigma_i^o + \Sigma_j^o
     -
     2\left((\Sigma_i^o)^{1/2} \Sigma_j^o (\Sigma_i^o)^{1/2}\right)^{1/2}
     \right].
\end{align}
\normalsize
The geometric cost $E_g^o(i, j)$ can be ambiguous in certain environments. To tackle this, we estimate a semantic neighbor histogram $H_i$ for object $\Ob_i$ that provides a topological overview of the local neighborhood graph of $\Ob_i$. The histogram encodes the distribution of semantic classes over $\mathcal{N}_i$, the objects adjacent to $\Ob_i$ in $\E^o$.
\begin{equation}\label{eq:st_hist}
    H_i = [\dots h_l \dots]^\top \in \R^{\abs{\mathcal{L}}},\ h_l = \sum_{j \in \mathcal{N}_i} \mathbb{I}(\objlabl_j = l), l \in \mathcal{L}.
\end{equation}
where $\mathcal{L}$ is the set of all class labels in $\G^o$, $\mathbb{I}(\cdot)$ is the indicator function. The structural cost $E_{st}^o(i, j)$ is then calculated on the basis of the L2 norm between the histograms of two objects
\begin{equation}
    E_{st}^o(i, j) = \norm{H_i - H_j}.
\end{equation}

The final cost $E^o(i,j)$ is a convex combination of geometric cost $E_g^o(i, j)$ and structural cost $E_{st}^o(i, j)$, controlled by a weighting coefficient $\alpha_c$

\begin{equation}\label{eq:cost}
    E^o(i,j) = \alpha_c \cdot E_g^o(i, j) + (1-\alpha_c) \cdot E_{st}^o(i, j).
\end{equation}

We build a cost matrix based on cost $E^o$ between all semantically valid pairs of nodes and solve for optimal assignment using the Hungarian algorithm~\cite{kuhn1955hungarian}. This provides a robust set of node-node correspondences $\C^o$ for subsequent graph registration.

\subsubsection{Coarse Graph Registration}

For the set of correspondences $\C^o$, we aim to estimate a rigid transformation $\Tf_\text{coarse} = (\rot, \trans)$, following the split pose representation~\cite{nguyen2024gptr}, that aligns the two graphs.
The fundamental challenge for graph registration solely on objects is that the sparse information it involves can lead to ambiguity in orientation, especially in scenes with few objects. We introduce a structure-aware registration approach which utilizes the graph nodes and the spatial relationship between objects in the graph. 
The cost function for alignment between the two object graphs $G_{t}^o$ and $\G_{t-1}^o$ can be defined as a combination of the node alignment cost $E_{\Ob}$ and the edge alignment cost $E_{\E}$:

\small
\begin{equation} \label{eq:co_reg}
    E(\Tf) = \sum_{(i,j) \in \mathcal{C}} E_{\Ob}(i,j) + \sum_{e \in \E} E_{\E}(e).
\end{equation}
\normalsize
where $\E^o_{\mathcal{C}}$ are the edges of $G^o_t$ whose two endpoints are both matched by $\C^o$, so that each has a counterpart in $\G^o_{t-1}$.
$E_\Ob$ is a probabilistic distribution-to-distribution cost in the manner of Generalized-ICP~\cite{segal2009generalized}, minimizing the Mahalanobis distance between two matched nodes under their combined covariance:

\begin{align}
    E_{\Ob}(i,j) = \norm{(\rot \mu_i^o + \trans) - \mu_j^o}^2_{W_\Ob},
    W_\Ob \triangleq \left[\Sigma_j^o + \rot \Sigma_i^o \rot^\top \right]^{-1}
\end{align}

To construct $E_\E$, for each edge connecting two nodes $u$ and $v$ in the graph, we take $K$ parameters $\gamma_1, \dots, \gamma_K$ evenly spaced in $(0,1)$ and interpolate the mean and covariance along the edge into intermediary Gaussians $(\mu_k^o, \Sigma_k^o)$:
\begin{align}
    \mu_k^o & = (1-\gamma_k) \mu_u^o + \gamma_k \mu_v^o, \space \\ \nonumber
    \Sigma^o_k & = (\Sigma^o_u)^{1/2} [(\Sigma^o_u)^{-1/2} \Sigma^o_v (\Sigma^o_u)^{-1/2}]^{\gamma_k} (\Sigma^o_u)^{1/2}.
\end{align}

This is the geodesic on $\spd$ under the Affine-Invariant Riemannian Metric rather than a Euclidean interpolation, and the difference is systematic rather than numerical. For two covariances related by a rotation, the same extent seen from two orientations, the flat midpoint satisfies $\det\!\big(\tfrac{1}{2}(\Sigma_u^o + \Sigma_v^o)\big) \ge \sqrt{\det \Sigma_u^o \, \det \Sigma_v^o}$, with equality only when the two agree. The determinant being the squared represented volume, the flat mean of two equal-volume objects is a strictly larger object than either. The geodesic keeps every virtual Gaussian on the manifold and at the volume its endpoints imply, throughout the tunnel from one node to the other.

This method produces a set of virtual edge Gaussian distributions $(\mu_{t-1,k}^o, \Sigma_{t-1,k}^o)$ for graph $\G_{t-1}^o$ and $(\mu_{t,k}^o, \Sigma_{t,k}^o)$ for graph $G_t^o$, which can be intuitively visualized as a ``tunnel'' that delineates a smooth transformation pathway from one Gaussian to another. 
In the same Mahalanobis form as $E_\Ob$, the constraints $E_\E$ are built on pairs of virtual Gaussians:

\small
\begin{equation}\label{eq:edge_cost}
\begin{aligned}
    E_{\E}(e)
    &= \sum_k
    \norm{
        (\rot \mu_{t,k}^o + \trans) - \mu_{t-1,k}^o
    }_{W_{\E,k}}^2, \\
    W_{\E,k}
    &\triangleq
    \left(
        \Sigma_{t-1,k}^o
        + \rot \Sigma_{t,k}^o \rot^\top
    \right)^{-1}.
\end{aligned}
\end{equation}
\normalsize

The transformation $\Tf_\text{coarse}$ is then estimated by solving the joint non-linear least squares problem:
\begin{equation}\label{eq:regis}
    \Tf_\text{coarse} = \argmin_{(\rot, \trans)} E(\Tf).
\end{equation}
For efficient solution, we initialize  $\Tf_\text{coarse}$ from the odometry estimate $\Tf_\text{init}$ of the motion between the last two frames (\Cref{fig:system}).
The coarse transformation estimate thus aligns both the HGG nodes and the spatial topology of the scene, and provides a robust initialization estimate for the complete registration problem.

\subsubsection{Fine Correspondence Estimation and Registration}\label{sec:fine_corr}

After $\mathcal{C}^o$ and $\Tf_\text{coarse}$ align the nodes at the object graph, local misalignment persists due to the sparse nature of the graphs. To further improve the result, we formulate a fine registration step that uses $\mathcal{C}^o$ and $\Tf_\text{coarse}$ as initialization.

Using a correspondence $(i,j)$ between objects $\Ob_i \in \V^o_t$ and $\Ob_j \in \V^o_{t-1}$, the objective is to find Gaussian-Gaussian correspondences for the geometry graphs $G_i^g$ and $\G_j^g$, as illustrated in \Cref{fig:corres}(c). 
Solely relying on geometric proximity for node association is challenging due to semantic and spatial ambiguities. To resolve this, we incorporate structural information using the Weisfeiler-Lehman (WL) framework~\cite{shervashidze2011weisfeiler}. While typically used for whole-graph comparison, we adapt the WL iteration to generate a structural feature vector $h_u$ for each individual node $u \in \V^g_i$. This allows us to define a structural distance metric $\mathcal{D}_{WL}(h_u, h_v)$ between two nodes $u$ and $v$ ($u\in\V_i^g$, $v\in\V_j^g$):
\begin{equation}
    \mathcal{D}_{WL}(h_u, h_v) = \norm{h_u - h_v}^2.
\end{equation}

The fine level matching cost $E^g$ is then determined by combining this feature with the spatially constrained distance
\begin{align}
    E^g(u, v)
    & = \alpha_r || \mu^{s}_u{'} - \mu^g_v ||^2
    + (1-\alpha_r) \mathcal{D}_{WL}(h_u, h_v),
\end{align}
where $\mu^g_u{'}$ is the centroid of a node in $G^g_i$ using the coarse estimate $\Tf_\text{coarse}$, which restricts the search space to the local neighborhood. Note that the Euclidean term here is the one place the pipeline compares geometry off the manifold, and the gating is what makes it admissible: it ranks candidates within a set the object graph stage has already selected under a distributional cost, so it cannot introduce a correspondence that a valid comparison would exclude. 

A cost matrix is constructed between all pairs of nodes between two graphs $G_i^g$ and $\G_j^g$, and solved for optimal assignment with the Hungarian algorithm~\cite{kuhn1955hungarian}, as shown in \Cref{fig:corres}(d). This method results in fine Gaussian-level correspondences for each geometry graph, which are aggregated into the correspondence set $\C^g$. Next, we estimate the refined pose $\hat{\Tf}_\text{fine}$ by solving \eqref{eq:regis} again, with $\C^g$ in place of $\C^o$ and the geometry graph edges in place of $\E^o_{\mathcal{C}}$.

\subsubsection{Map Update}

Map update incorporates the local graph into the global scene graph using the refined pose $\Tf_\text{fine}$ and node correspondences $\C^o$. 
The object nodes associated through $\mathcal{C}^o$ are merged in the global scene graph. Because each node carries its evidence count, the merge is no longer a symmetric combination of two equally trusted estimates but a fusion weighted by accumulated evidence. 
For an associated pair $(i,j)$, the node from the local graph $G^o_t$ is first brought into the global scene graph $\G_{t-1}^o$, $\mu_i^o \leftarrow \rot \mu_i^o + \trans$ and $\Sigma_i^o \leftarrow \rot \Sigma_i^o \rot^\top$, after which the two states $(\mu_i, \Sigma_i, \kappa_i)$ and $(\mu_j, \Sigma_j, \kappa_j)$ merge into:
\begin{equation}\label{eq:merge}
\begin{aligned}
    \kappa &= \kappa_i + \kappa_j, \quad
    \mu = \frac{\kappa_i\,\mu_i + \kappa_j\,\mu_j}{\kappa}, \\
    \Sigma &= \frac{1}{\kappa}
    \left(\kappa_i\Sigma_i + \kappa_j\Sigma_j\right)
    + \frac{\kappa_i\,\kappa_j}{\kappa^2}
    \big(\mu_i - \mu_j\big)\big(\mu_i - \mu_j\big)^\top.
\end{aligned}
\end{equation}
The merge accumulates disjoint evidence into a single belief by adding raw statistics rather than applying any distance metric, and is exactly the conjugate update of \Cref{eq:niw}. The counts add, $\kappa = \kappa_i+\kappa_j$, so the calibration contracts rather than being preserved as an average would, and the between-means term composes two partial views rather than swelling one extent. Nodes of $G^o_t$ left unmatched by $\C^o$ enter the global scene graph $\G_t$ as new objects with their counts unchanged. Similarly, we merge the geometry graphs using the geometry-graph correspondences $\C^g$.

Nodes of stuff classes such as road, building, and terrain are unbounded along the traversal, so merging them by \eqref{eq:merge} alone would accumulate a single node whose mean and covariance describe no physical extent. Therefore for nodes of classes, we admit a merge only when the two nodes are within a spatial bound, and instantiate a new node otherwise, so that such nodes are carried as a sequence of bounded submaps rather than one growing node.

\subsection{Correction}\label{sec:Optimization}
Local correction operates on a sliding window of the $M$ most recent frames, addressing two errors with different scopes: pose drift induces a correlated displacement of all nodes observed within the window, whereas an error in an individual object's extent is local and can only be corrected through further observations of that object. Accordingly, correction jointly refines the representation at two levels: a factor graph optimizes the frame poses $\mathcal{X}$ and the object frames $\theta_i^o=(\mu_i^o,\Sigma_i^o)$ (\Cref{sec:fgo}), while an EM step refines the frame-conditioned internal geometry $\theta_i^g$ of each geometry graph $G_i^g$ in its object frame (\Cref{sec:em}). The two optimizations are nested: the factor graph does not update the geometry graphs, and EM does not update the poses or object frames.

\subsubsection{Factor Graph Optimization}\label{sec:fgo}
The variables are the poses in the sliding window, $\mathcal{X} = \{ \Tf_1, \Tf_2, \dots, \Tf_M \}$, obtained from alignment, and the frames $\theta^{o}_i = (\mu^o_i, \Sigma^o_i)$ of the objects $\Ob_i$ observed within it. The internal geometry of each object is held fixed at this level. Poses and frames are estimated in one graph because their Fisher information blocks are strongly coupled: an object's frame is observed only through the poses it was seen from, so neither resolves without the other. The cost function is:

\small
\begin{equation}\label{eq:fgo}
\begin{aligned}
E(\mathcal{X}, \{\theta_i^o\})
&=
\sum_{m=2}^{M}
\underbrace{
    \left\|
        \mathbf{r}_{\mathrm{odom}}
        \left(\hat{\Tf}_{m-1}, \hat{\Tf}_m\right)
    \right\|_{\Omega_{\mathrm{odom}}^m}^{2}
}_{\text{Odometry Constraints}}
\\
&\quad+
\sum_{(m,i)\in\mathcal{F}}
\underbrace{
    \left\|
        \mathbf{r}_{\mathrm{obj}}
        \left(\hat{\Tf}_m, \Ob_i\right)
    \right\|_{\Omega_{\mathrm{obj}}^{(m,i)}}^{2}
}_{\text{Object Constraints}}.
\end{aligned}
\end{equation}
\normalsize
where $\Omega_{\mathrm{odom}}^m$ is the information matrix associated with the odometry measurement at frame $m$, and $\Omega_{\mathrm{obj}}^{(m,i)}$ is the information matrix associated with the observation of object $\Ob_i$ from frame $\Tf_m$, given by
\begin{equation*}
\Omega_{\mathrm{obj}}^{(m,i)}
=
\breve{\kappa}_{m,i}^o
(\breve{\Sigma}_{m,i}^o)^{-1}
\oplus
\frac{\breve{\kappa}_{m,i}^o}{2}\mathbf{I}_6.
\end{equation*}
Here, $\mathcal{F}$ is the set of object factors, with one factor for each pair $(m,i)$ such that $\Ob_i$ is observed from $\Tf_m$. Each observation consists of a mean $\breve{\mu}_{m,i}^o$, a covariance $\breve{\Sigma}_{m,i}^o$, and a point count $\breve{\kappa}_{m,i}^o$. The point count determines the strength of the corresponding constraint: an object supported by more measurements contributes greater information than one observed only sparsely. Importantly, $\breve{\kappa}_{m,i}^o$ denotes the evidence contributed by the observation from frame $m$, rather than the accumulated count $\kappa_i^o$ maintained by the corresponding object in the global graph. The odometry factor constrains consecutive frame poses according to the measured relative transformation $\breve{\Tf}^{m-1}_{m}$:
\begin{equation}\label{eq:odom_res}
    \textbf{r}_\text{odom}\left( \hat{\Tf}_{m-1}, \hat{\Tf}_m \right) = \hat{\Tf}_{m-1}^{-1} \hat{\Tf}_{m}\boxminus\breve{\Tf}^{m-1}_{m}.
\end{equation}
The object factor constrains a pose against the objects observed from it. The parameters of $\Ob_i$ are first brought into the body frame of $\Tf_m = (\rot_m, \trans_m)$,
\begin{equation}\label{eq:trans}
    \mu'^{o}_{m,i} = \rot_m^\top (\mu_i^o - \trans_m),\ 
    \Sigma'^{o}_{m,i}  = \rot_m^\top \Sigma_i^o \rot_m,
\end{equation}
and compared with the observation:

\small
\begin{align}\label{eq:obj_res}
    \textbf{r}_\text{obj}\left(\cdot\right)
    = 
    \begin{bmatrix}
        \textbf{r}_{\mu} \\
        \textbf{r}_{\Sigma}
    \end{bmatrix}
   =
    \begin{bmatrix}
        \mu'^{o}_{m,i} - \breve{\mu}_{m,i}^{o} \\
        \mathrm{vech}_W\big(U^{\top}(\log\Sigma'^{o}_{m,i} - \log\breve{\Sigma}^{o}_{m,i})\, U \big)
    \end{bmatrix},
\end{align}
\normalsize
where $\breve{\Sigma}^{o}_{m,i} = U\,\mathrm{diag}(\lambda_1,\lambda_2,\lambda_3)\,U^{\top}$ is the eigendecomposition of the observed covariance, computed once per factor, where $\lambda_*$ is the extent along the corresponding principal axis. Working in this basis, the residual compares the two covariances axis by axis: $a$ and $b$ index the axes; $\delta_{ab} = \log\lambda_a - \log\lambda_b$ is the gap between two of them, and $\mathrm{vech}_W$ is the half-vectorization~\cite{pennec2006riemannian, higham2008functions} with entry $(a,b)$ scaled by:

\small
\begin{equation}
    w_{ab} = 
    \begin{cases}
    1, & a=b, \\[2pt]
    \sqrt{2}\,\dfrac{\sinh(\delta_{ab}/2)}{\delta_{ab}/2}, & a<b.
    \end{cases}
\end{equation}
\normalsize
The weights $w_{ab}$ make $\mathrm{vech}_W$ an isometry from the tangent space at $\breve{\Sigma}^{o}_{m,i}$ under the affine-invariant metric onto Euclidean $\mathbb{R}^6$. The resulting residual $\mathbf{r}_{\Sigma}$ therefore captures all six degrees of freedom of $\Sigma_i^o$, allowing the factor-graph optimization to jointly refine the object's spatial extent and principal-axis orientation while remaining consistent with the geometry of $\spd$. Finer geometric structure beyond the single Gaussian is represented and refined by the geometry graph.

\subsubsection{EM Optimization}\label{sec:em}
A topology-constrained EM refines the nodes of each $G^g_i$ in its object frame while keeping $\E^g_i$ fixed. The objects are refined independently, as permitted by the Fisher information: conditioned on their object frames, the internal geometries of distinct objects have no cross-information.
\paragraph{E-Step}
We estimate the probability that each observed point belongs to a specific Gaussian node. Let $\mathcal{P}_i^{\mathcal{W}} = \{\pos_l\}_{l=1}^{L_i}$ denote the points attributed to $\mathcal{O}_i$ within the sliding window, where $L_i=|\mathcal{P}_i|$, and let $k = 1,\dots,N_i$ index the nodes of $G^g_i$ with mixing weights $\pi_k={\kappa_k^g}/{\sum_{r=1}^{N_i}\kappa_r^g}$ set by their evidence counts $\kappa_k^g$. Given the node parameters $(\mu_k, \Sigma_k)$, the responsibility $\gamma_{l,k}$ of point $\pos_l$ with respect to node $k$ is
\begin{equation}
    \gamma_{l,k}^{(t)} =
    \frac{\pi_k^{(t)}\,
          \mathcal{N}\!\left(\pos_l | \mu_k^{(t)}, \Sigma_k^{(t)}\right)}
         {\sum_{m=1}^{N_i} \pi_m^{(t)}\,
          \mathcal{N}\!\left(\pos_l | \mu_m^{(t)}, \Sigma_m^{(t)}\right)},
    \label{eq:e-step}
\end{equation}
so that $\sum_{k=1}^{N_i} \gamma_{l,k}^{(t)} = 1$ for every point.
\paragraph{M-Step}
We update the Gaussian parameters of the geometry graph by maximizing the expected complete-data log-likelihood, equivalently minimizing the KL divergence to the empirical point distribution. Writing $n_k$ for the responsibility mass accumulated by node $k$, which is the re-estimated count $\kappa^g_k$ of that node, the parameters are re-estimated as:
\begin{align}
    n_k &= \sum_{l=1}^{L_i} \gamma_{l,k}^{(t)},
    \quad \hat{\pi}_k = \frac{n_k}{L_i},
    \quad \hat{\mu}_k = \frac{1}{n_k}
        \sum_{l=1}^{L_i} \gamma_{l,k}^{(t)}\, \pos_l,
    \label{eq:m-step-mu} \\[2pt]
    \hat{\Sigma}_k &= \frac{1}{n_k} \sum_{l=1}^{L_i} \gamma_{l,k}^{(t)}
        \left(\pos_l - \hat{\mu}_k\right)
        \left(\pos_l - \hat{\mu}_k\right)^{\!\top}
        + \lambda \mathbf{I}_3,
    \label{eq:m-step-sigma}
\end{align}

Iterating \eqref{eq:e-step}--\eqref{eq:m-step-sigma} lets the geometry graph morph toward finer geometric detail while the edge set $\mathcal{E}^g_i$ is held fixed: only the node parameters and their counts are re-estimated, so the topology built at creation persists through refinement and continues to relate the parts of the object to one another.

\begin{remark}\label{rem:principle}
The two levels could in principle be jointly optimized, but the object-frame formulation makes their coupling negligible at the resolution of the internal geometry. Under the gauge constraints of the supplementary material~\cite{supplementary}, all cross-information between pose and internal geometry vanishes except for a channel coupling rotation and covariance, whose normalized magnitude is $O(\varepsilon)$, where $\varepsilon=s/\rho$ is the ratio of internal scale to object extent. Consequently, decoupling the factor-graph and EM optimizations incurs only an $O(\varepsilon^2)$ difference in the Cram\'er--Rao bound, as proved in the supplementary material~\cite{supplementary}. Under model or attribution errors, the same coupling can instead propagate errors between pose and internal geometry; the decoupled formulation removes this transmission path.
\end{remark}

\subsubsection{Loop Closure and Pose Graph Optimization}\label{sec:loop}
Global correction reuses the representation rather than adding to it: loop closure candidates are retrieved from the graph's own topology, verified by the coarse-to-fine alignment of \Cref{sec:Alignment} applied unchanged, and the accepted constraints enter a standard pose graph optimization. Needing no loop-specific descriptor and no dense re-registration is a property of the representation rather than a separate contribution; the algorithm and its thresholds are given in the supplementary material~\cite{supplementary}.


\section{Experiment}\label{sec:experiments}
We evaluate HGG in three parts, across indoor, outdoor, and cross-modality data. The main benchmark measures the complete system against representative scene graph and object-SLAM systems on object accuracy and computational cost. The module evaluation then tests creation, alignment, and correction separately, each against the claim it was designed to support. The downstream evaluation asks whether the resulting graph is actionable.

\subsection{Experiment Setup}\label{sec:setup}

\begin{table}[t]
\centering
\caption{Evaluation Datasets}
\label{tab:dataset}
\def\arraystretch{1.15}
\setlength{\tabcolsep}{1.1pt}
\begin{tabular}{llll}
\toprule
Dataset     & Sensor           & Environment    & Semantic Labels   \\ \midrule
uHumans2    & RGB-D            & indoor         & Simulator~\cite{hughes2022hydra}      \\
KITTI       & LiDAR         & outdoor        & RangeNet++~\cite{milioto2019iros}                 \\
KITTI-Carla & LiDAR & outdoor        & Simulator~\cite{deschaud2021kitti}         \\
MCD         & LiDAR         & outdoor        & 2DPASS\cite{yan20222dpass}            \\
ScanNet-SG  & RGB-D            & indoor         & ScanNet-SG~\cite{scannet_sg}                 \\
In-house    & LiDAR+RGB   & indoor+outdoor & Mask2Former\cite{cheng2021mask2former}+SAM2\cite{ravi2024sam2}  \\
\bottomrule
\end{tabular}
\vspace{-0.5cm}
\end{table}

\subsubsection{Datasets}

\Cref{tab:dataset} summarizes the six datasets used in our evaluation. uHumans2~\cite{rosinol2021kimera} provides three structurally distinct indoor environments (Apartment, Office, Subway). The outdoor evaluation spans three LiDAR datasets of increasing difficulty: KITTI~\cite{geiger2013vision}; KITTI-Carla~\cite{deschaud2021kitti}, whose noise-free simulated ground truth isolates odometry failures as the sole source of localization error in this demanding regime; and MCD~\cite{nguyen2024mcd}, the hardest case, where predicted labels are evaluated against survey-grade Leica RTC360 reference maps, so the geometry is exact but the semantics are not. We use all KITTI sequences except 01 and 04 (highway scenes with too few objects to populate a graph), all seven KITTI-Carla sequences, and all six sequences from each of MCD's NTU and KTH campuses. Results reported for KITTI, KITTI-Carla, MCD-NTU, and MCD-KTH are averaged over their sequences. ScanNet-SG~\cite{scannet_sg} poses an offline pairwise alignment task: recover the rigid transform between two independently segmented submaps from adjacent ScanNet sequences, with no temporal dependence on the originating trajectory. We use the Subscan test split (Scenes 600--705, 3,435 pairs), whose ground-truth correspondences are defined by 3D instance IoU rather than semantic consistency, leaving 27.4\% of pairs inter-class. Two in-house sequences test deployment beyond controlled benchmarks, collected with the Manifold Odin suite~\cite{odin1}: a solid-state LiDAR with a $120^{\circ}$ horizontal FoV, an RGB camera, and a built-in visual-inertial-LiDAR odometry pipeline whose output serves as reference poses. The first covers multiple floors of a lab building; the second combines indoor and outdoor traversal.

\subsubsection{Ground Truth Scene Graph}
For each dataset, we generate a ground-truth scene graph directly from the ground-truth poses, accumulating the raw semantic point clouds and partitioning them into object instances. The partition follows the procedure used for the instance annotations of SemanticKITTI~\cite{behley2019semantickitti}: semi-automated clustering with per-class parameters, then manual inspection and revision against a single criterion: each segment corresponds to one physically distinct object, and each object to one segment. This differs from the protocol of~\cite{hughes2022hydra}, which clusters a batch mesh reconstruction: accumulating points under known poses introduces fewer processing stages between sensor and reference, whereas a mesh carries its own surface-fitting error and imposes the object boundaries the reconstruction produced. Absolute values here are consequently not comparable with those reported there. On MCD, the question does not arise, as the reference is anchored to a survey-grade Leica RTC360 scan.

\subsubsection{Metrics}\label{sec:metrics}
The object accuracy metric follows~\cite{hughes2022hydra}: each predicted node is matched to the nearest unmatched ground-truth instance within a fixed radius, with matches counted as true positives and unmatched predictions as false positives. We report the position error $\Delta_p$, the mean centroid displacement of matched pairs in meters, together with precision, recall, and F1. The radius is set per dataset to the scale of the odometry error, adopting the values of~\cite{hughes2022hydra} on uHumans2; all values are listed in the supplementary material~\cite{supplementary}. Graph alignment follows the ScanNet-SG protocol~\cite{chen2026opensga}, reporting accuracy, precision, recall, and F1 over correspondences at the full-set level. Trajectory accuracy is the absolute trajectory error (ATE), and geometric fidelity is quantified by the Chamfer distance and map accuracy~\cite{hu2024paloc, hu2025mapeval}, both computed between the points extracted from the combined geometry graphs and the survey-grade reference, in meters. Computational cost is the per-frame processing time and the RAM footprint in megabytes as a function of sequence length, both measured on a desktop (i7-12700 CPU with 128 GB RAM and RTX 3080 GPU). 

\subsubsection{Baselines}
Indoors we compare against three systems that derive their scene graph from a dense reconstruction: Kimera~\cite{rosinol2021kimera}, Hydra~\cite{hughes2022hydra}, which maintains a hierarchical metric--semantic map, and ConceptGraphs~\cite{gu2024conceptgraphs}, which is open-vocabulary. We also compare against the object-SLAM systems OA-SLAM~\cite{zins2022oa} and VOOM~\cite{wang2024voom}, which represent objects as deterministic quadrics. Outdoors, object-level 3D scene graph systems are effectively absent, as the systems above target structured indoor RGB-D, so we adopt SG-SLAM~\cite{wang2025leveraging}, which maintains a global object-level semantic graph.

All indoor systems receive the same Kimera-VIO odometry, except OA-SLAM and VOOM, which require an ORB-SLAM2 front-end and are therefore compared against HGG run on that front-end. For outdoor experiments, both SG-SLAM and HGG use KISS-ICP\cite{vizzo2023kiss} as the front-end odometry. For loop closure, Kimera, Hydra, and SG-SLAM close their own loops; ConceptGraphs uses the loop-closed trajectory our system produces. Graph alignment baselines are taken from~\cite{chen2026opensga} and are not re-run.

\subsection{Main Benchmark}

\subsubsection{Object Accuracy Evaluation}

We follow the thing/stuff distinction of SemanticKITTI~\cite{behley2019semantickitti} and evaluate over thing classes only, since stuff classes such as road, building, vegetation, and terrain admit no well-posed decomposition into instances. Stuff nodes are still built and still carry the spatial relations the graph depends on; however, they are not considered in the object accuracy evaluation. On uHumans2, the split follows the label space of~\cite{hughes2022hydra}. \Cref{tab:obj_eval} reports the complete evaluation.

Indoors, all methods use the same Kimera-VIO providing accurate odometry input, so differences stem from how each system converts observations into map nodes rather than from localization. The baselines cluster at low F1, Kimera and Hydra between 0.26 and 0.50 and ConceptGraph between 0.09 and 0.33, each limited by recall. In contrast, HGG reaches F1 0.83$\sim$0.88 at recall 0.79$\sim$0.97, recovering the large majority of ground-truth objects. Two mechanisms separate HGG from the baselines. A node collapsed to a point at creation cannot be distinguished from its neighbours once drift displaces it, so two views of one object either merge into a ghost or fail to associate, whereas a node carrying its covariance stays associable through a distributional distance. Independently of drift, a graph derived from a dense reconstruction can contain only the objects that the reconstruction builds, a ceiling on recall that better poses do not lift (\Cref{sec:odom_rob}). Subway shows both at once: Hydra holds precision 0.93 at recall 0.34, admitting only objects confirmed from many views.

The two in-house sequences test deployment outside a controlled benchmark. The platform is LiDAR-visual-inertial, the semantic labels are predicted rather than clean, and the sequence In-house-02 traverses from an indoor corridor into open outdoor space, enlarging the mapped volume by orders of magnitude mid-sequence. 
The baselines degrade sharply under that pressure, to F1 between 0.17 and 0.60. Cost is part of it, but the sharper penalty is due to determinism: with predicted rather than clean labels, a node fused from a mislabeled or partial first view carries no record of how little evidence stood behind it, so later, better views cannot outweigh it.
HGG holds at F1 0.94 indoors and 0.88 across the transition, at real-time rates. Two properties account for it. Geometry lives in the node, so no dense map is built first and nothing scales with the volume traversed. Moreover, a node's belief is set by the evidence behind it, so an object seen partially, which a $120^{\circ}$ field of view makes common, stays associable instead of being committed to the shape of its first view; \Cref{sec:niw_abl} ablates exactly this on In-house-02.

The sensor changes outdoors, the scale grows, and the data gets sparser, across four settings and three datasets. These scenes are where deterministic association is most exposed, being dominated by repeated, weakly distinctive structure: a point-based semantic graph facing many near-identical candidates, with no distributional distance, resolves them inconsistently and drops objects. SG-SLAM reports F1 between 0.54 and 0.67, again limited by recall, while HGG reaches 0.80$\sim$0.91 at recall 0.79$\sim$0.93 by disambiguating those candidates through their full distributional geometry rather than centroid proximity. MCD is the sharpest case since its geometry is measured by a terrestrial laser scan while the semantics feeding the system are noisy predictions. HGG attains the best position error and F1 above 0.80 on both campuses, at \SIrange{14}{20}{\milli\second} per frame.


\begin{figure*}
    \centering
    \includegraphics[width=0.95\linewidth]{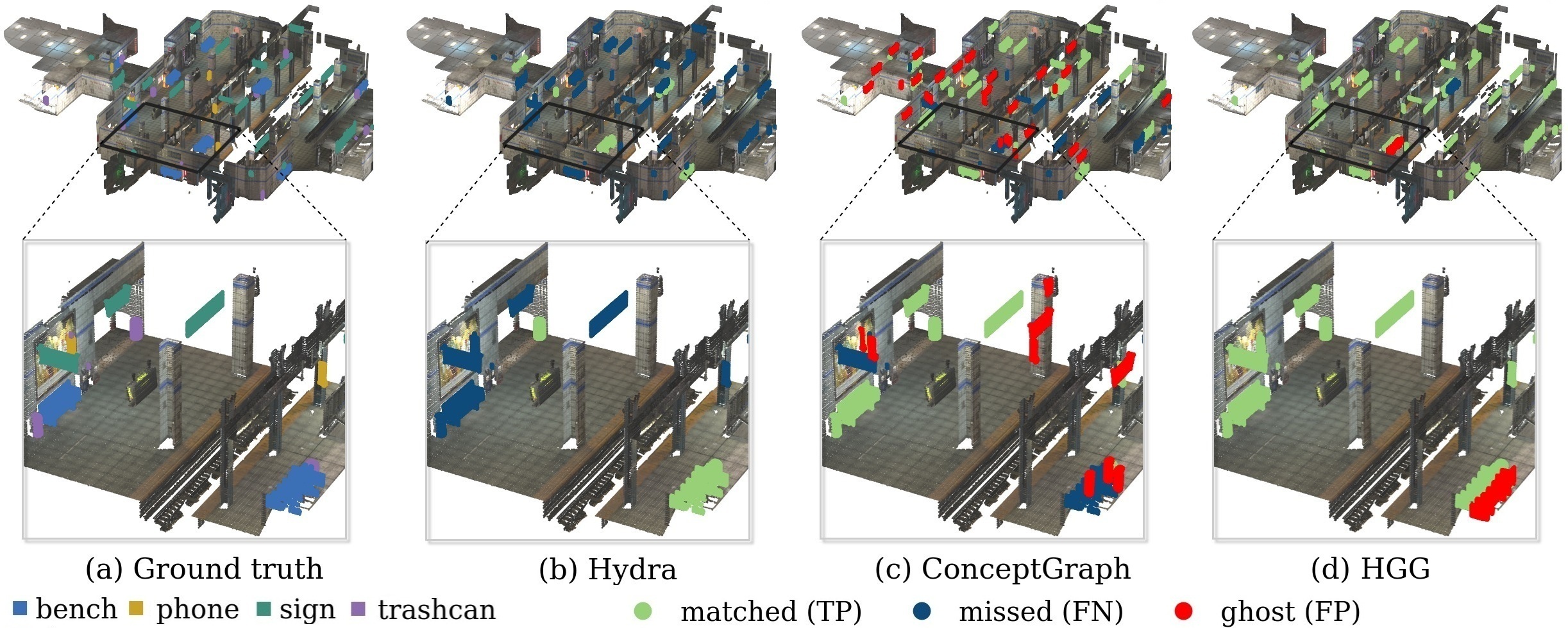}
    \caption{Qualitative object accuracy on uHumans2-Subway. All panels share Ground-truth rgb point cloud map and objects colored by classes in (a), and matching accuracy in (b-d). Hydra~(b) is dominated by missed detection (blue), since a graph abstracted from a dense reconstruction can contain only what that reconstruction builds. ConceptGraph~(c) reaches most instances but re-instantiates them, surrounding recovered objects with ghosts (red). HGG~(d) cover almost all object nodes with high precision.}
    \label{fig:qualitative}
    \vspace{-0.2cm}
\end{figure*}

\begin{table*}[t]
\centering
\caption{Object Accuracy Evaluation}
\def\arraystretch{0.8}
\label{tab:obj_eval}
{\footnotesize
\begin{tabularx}{\textwidth}{lYYYYrYYYYrYYYYr}
\toprule
& \multicolumn{5}{c}{Apartment} & \multicolumn{5}{c}{Office} & \multicolumn{5}{c}{Subway} \\
\cmidrule(lr){2-6}\cmidrule(lr){7-11}\cmidrule(lr){12-16}
\textbf{Method} & $\Delta_p$ & Recall & Prec. & F1 & t (ms) & $\Delta_p$ & Recall & Prec. & F1 & t (ms) & $\Delta_p$ & Recall & Prec. & F1 & t (ms) \\
\midrule
Kimera       & 0.13 & 0.30 & 0.51 & 0.38 & 2018.5 & 0.26 & 0.29 & 0.50 & 0.36 & 1956.7 & 0.73 & 0.34 & 0.78 & 0.47 & 3742.8  \\
Hydra        & 0.14 & 0.26 & 0.58 & 0.35 & 62.8 & 0.18 & 0.18 & 0.49 & 0.26 & 58.4 & 1.46 & 0.34 & \textbf{0.93} & 0.50 & 41.6  \\
ConceptGraph & 0.21 & 0.23 & 0.77 & 0.36 & 1033.8 & 0.26 & 0.24 & 0.47 & 0.32 & 3169.1 & 0.82 & 0.65 & 0.23 & 0.34 & 4672.7  \\
HGG       & \textbf{0.06} & \textbf{0.79} & \textbf{0.94} & \textbf{0.86} & \textbf{29.3} & \textbf{0.11} & \textbf{0.79} & \textbf{0.86} & \textbf{0.83} & \textbf{28.5} & \textbf{0.55} & \textbf{0.97} & 0.80 & \textbf{0.88} & \textbf{30.6}  \\
\end{tabularx}
}
\vspace{2.5pt}
{\footnotesize
\begin{tabularx}{\textwidth}{lYYYYRYYYYR}
\toprule
& \multicolumn{5}{c}{In-house-01} & \multicolumn{5}{c}{In-house-02} \\
\cmidrule(lr){2-6}\cmidrule(lr){7-11}
\textbf{Method} & $\Delta_p$ & Recall & Precision & F1 & t (ms) & $\Delta_p$ & Recall & Precision & F1 & t (ms) \\
\midrule
Kimera       & 0.21 & 0.47 & 0.70 & 0.56 & 699.8 & 0.33 & 0.51 & 0.36 & 0.43 & 1649.4  \\
Hydra        & 0.25 & 0.52 & 0.70 & 0.60 & 68.5 & 0.10 & 0.21 & 0.38 & 0.25 & 61.8  \\
ConceptGraph & 0.73 & 0.25 & 0.13 & 0.17 & 2046.6 & 0.43 & 0.22 & 0.23 & 0.22 & 3641.0  \\
HGG       & \textbf{0.03} & \textbf{0.88} & \textbf{1.00} & \textbf{0.94} & \textbf{43.6} & \textbf{0.05} & \textbf{0.92} & \textbf{0.85} & \textbf{0.88} & \textbf{50.6} \\
\end{tabularx}
}
\vspace{2.0pt}
{\footnotesize
\begin{tabularx}{\textwidth}{lYYYYYYYYYYYYYYYYYYYYYYYYY}
\toprule
& \multicolumn{5}{c}{KITTI-Carla} & \multicolumn{5}{c}{KITTI} & \multicolumn{5}{c}{MCD-KTH} & \multicolumn{5}{c}{MCD-NTU} \\
\cmidrule(lr){2-6}\cmidrule(lr){7-11}\cmidrule(lr){12-16}\cmidrule(lr){17-21}
\textbf{Method} & $\Delta_p$ & Recall & Prec. & F1 & t(ms) & $\Delta_p$ & Recall & Prec. & F1 & t(ms) & $\Delta_p$ & Recall & Prec. & F1 & t(ms) & $\Delta_p$ & Recall & Prec. & F1 & t(ms) \\
\midrule
SG-SLAM & 1.18 & 0.63 & 0.62 & 0.63 & 23.5 & \textbf{1.22} & 0.75 & 0.61 & 0.67 & \textbf{14.2} & 0.58 & 0.65 & 0.46 & 0.54 & 40.3 & 0.86 & 0.78 & 0.45 & 0.56 & 51.0  \\
HGG  & 1.12 & \textbf{0.89} & \textbf{0.88} & \textbf{0.88} & \textbf{13.4} & 1.26 & \textbf{0.93} & \textbf{0.90} & \textbf{0.91} & 26.1 & \textbf{0.52} & \textbf{0.79} & \textbf{0.83} & \textbf{0.80} & 14.4 & \textbf{0.75} & \textbf{0.84} & \textbf{0.82} & \textbf{0.82} & \textbf{20.1} \\
\bottomrule
\end{tabularx}
}
\begin{tablenotes}
\footnotesize
\item $\Delta_p$: Position error (m), t: per-frame runtime in milliseconds.
\end{tablenotes}
\vspace{-0.2cm}
\end{table*}


\subsubsection{Object Accuracy under Varying Odometry}\label{sec:odom_rob}


\begin{table*}[!htb]
\centering
\caption{Object Accuracy Evaluation with Ground-Truth Poses and Different Odometry.}
\def\arraystretch{0.8}
\small
\label{tab:obj_eval_gt}

{\footnotesize
\begin{tabularx}{\textwidth}{lYYYYYYYYYYYYYYYY}
\toprule
& \multicolumn{4}{c}{Apartment} & \multicolumn{4}{c}{Office} & \multicolumn{4}{c}{Subway} \\
\cmidrule(lr){2-5}\cmidrule(lr){6-9}\cmidrule(lr){10-13}
\textbf{Method} & $\Delta_p$ & Recall & Prec. & F1 & $\Delta_p$ & Recall & Prec. & F1 & $\Delta_p$ & Recall & Prec. & F1 \\
\midrule
Hydra+GT        & 0.12 & 0.36 & 0.76 & 0.50 & 0.14 & 0.33 & 0.89 & 0.48 & 0.60 & 0.37 & \textbf{1.00} & 0.54  \\
ConceptGraph+GT & 0.06 & 0.15 & 0.62 & 0.24 & 0.12 & 0.20 & 0.57 & 0.30 & 0.40 & 0.62 & 0.37 & 0.46  \\
HGG+GT       & \textbf{0.04} & \textbf{0.87} & \textbf{0.98} & \textbf{0.92} & \textbf{0.05} & \textbf{0.88} & \textbf{0.90} & \textbf{0.89} & \textbf{0.10} & \textbf{0.98} & 0.97 & \textbf{0.97}  \\
\end{tabularx}}


{\footnotesize
\begin{tabularx}{\textwidth}{lYYYYYYYYYYYYYYYYYYYY}
\toprule
& \multicolumn{4}{c}{KITTI-Carla} & \multicolumn{4}{c}{KITTI} & \multicolumn{4}{c}{MCD-KTH} & \multicolumn{4}{c}{MCD-NTU}  \\
\cmidrule(lr){2-5}\cmidrule(lr){6-9}\cmidrule(lr){10-13}\cmidrule(lr){14-17}
\textbf{Method} & $\Delta_p$ & Recall & Prec. & F1 & $\Delta_p$ & Recall & Prec. & F1 & $\Delta_p$ & Recall & Prec. & F1 & $\Delta_p$ & Recall & Prec. & F1 \\
\midrule
HGG+GT Odom & \textbf{0.26} & 0.87 & \textbf{0.98} & \textbf{0.92} & \textbf{0.21} & 0.92 & \textbf{0.95} & \textbf{0.94} & \textbf{0.25} & \textbf{0.80} & \textbf{0.84} & \textbf{0.82} & \textbf{0.37} & \textbf{0.83} & \textbf{0.87} & \textbf{0.84}  \\
HGG+GICP    & 1.13 & \textbf{0.88} & 0.84 & 0.86 & 1.57 & 0.92 & 0.89 & 0.90 & 0.66 & 0.79 & 0.81 & 0.80 & 1.35 & 0.81 & 0.80 & 0.81  \\
HGG+FLOAM   & 1.13 & 0.87 & 0.90 & 0.88 & 1.13 & \textbf{0.93} & 0.91 & 0.92 & 0.73 & 0.78 & 0.81 & 0.79 & 1.22 & 0.83 & 0.81 & 0.81   \\
\bottomrule
\end{tabularx}
}

\vspace{-0.5cm}
\end{table*}


To separate representational quality from localization error, we re-run the indoor comparison with ground-truth poses. HGG improves F1 from 0.83$\sim$0.88 to 0.89$\sim$0.97. The baselines improve too but remain far below it, Hydra recovering at most 0.37 of ground-truth objects and ConceptGraph at most 0.62, so their low recall is not a localization artifact that better poses can fix. A graph read off a dense reconstruction can contain only what the reconstruction builds. That imposes a floor: the mesh or TSDF is held at a fixed voxel resolution, typically 10–20 cm, so a smaller object never reaches the graph however accurate the poses are. Creating nodes directly from the segmented point cloud replaces that floor with the sensor's own sampling density, so what can become a node is limited by what was measured rather than by what the map stores.

Outdoors we vary the odometry source instead, running HGG with ground-truth odometry, GICP, and FLOAM on the same sequences. F1 stays between 0.79 and 0.94 across all three, and recall is almost unchanged, within 0.02 of the ground-truth-odometry value on every dataset, while the position error moves with odometry quality. Alignment and correction therefore absorb the difference between front-ends rather than passing it to the map, so recovering the objects does not depend on having a strong odometry front-end available.


\subsubsection{Run-time Evaluation}
A representation whose cost scales with the number of objects rather than with the volume of the environment should hold its per-frame cost at sensor rate regardless of scene scale. The proposed system runs at \SIrange{28.5}{30.6}{\milli\second} on indoor RGB-D, \SIrange{43.6}{50.6}{\milli\second} on the in-house solid-state LiDAR, and \SIrange{13.4}{26.1}{\milli\second} on 64- and 128-channel outdoor LiDAR (\Cref{tab:obj_eval}), without reconfiguration between them and at or above the acquisition rate of every sensor in the evaluation. The gap to the dense baselines is structural: Kimera and ConceptGraphs need \SIrange{0.7}{4.7}{\second} per frame indoors and in-house, one to two orders of magnitude slower, because each frame must be fused into a mesh or point cloud before the semantic layer can update. Hydra, built for real-time operation, narrows this to \SIrange{41.6}{68.5}{\milli\second}, but cannot go below the cost of maintaining its ESDF and mesh layers. For HGG, the per-frame cost is dominated by front-end clustering, while creation and alignment scale with the number of object nodes in the current window, a quantity that varies far less across sensors and environments than the raw point count or the mapped volume.

Memory behaves the same way and is reported in the supplementary material~\cite{supplementary}: our system stays below \SI{1}{\giga\byte} on every sequence and plateaus once the object population of the traversed area is instantiated, whereas the dense baselines grow with trajectory length, exceeding \SI{50}{\giga\byte} on In-house-02.


\subsubsection{Object Accuracy against Object SLAM}

\begin{table*}[!htb]
\centering
\caption{Object Accuracy Evaluation Against Object SLAM on Indoor Datasets}
\setlength{\tabcolsep}{7.5pt}
\label{tab:obj_eval_ObSLAM}
{\footnotesize
\begin{tabular}{lrrrrrrrrrrrr}
\toprule
& \multicolumn{4}{c}{\textbf{Apartment}}                                              & \multicolumn{4}{c}{\textbf{Office}}                                                 & \multicolumn{4}{c}{\textbf{Subway}}                                                 \\ \cmidrule(l){2-5} \cmidrule(l){6-9} \cmidrule(l){10-13} 
& \multicolumn{1}{c}{$\Delta_p$} & \multicolumn{1}{c}{Recall} & \multicolumn{1}{c}{Precision} & \multicolumn{1}{l}{F1-score} & \multicolumn{1}{c}{$\Delta_p$} & \multicolumn{1}{c}{Recall} & \multicolumn{1}{c}{Precision} & \multicolumn{1}{l}{F1-score} & \multicolumn{1}{c}{$\Delta_p$} & \multicolumn{1}{c}{Recall} & \multicolumn{1}{c}{Precision} & \multicolumn{1}{l}{F1-score} \\ \midrule
OA-SLAM & 0.23                                                                              & 0.17                       & 0.33                          & 0.23                         & 0.46                                                                              & 0.33                       & 0.50                          & 0.40                         & 1.53                                                                              & 0.10                       & 1.00                          & 0.19                         \\
VOOM    & 0.41                                                                              & 0.18                       & 0.27                          & 0.21                         & 0.60                                                                              & 0.30                       & 0.38                          & 0.33                         & 0.76                                                                              & 0.05                       & 0.13                          & 0.08                         \\
HGG  & \textbf{0.06}                                                                     & \textbf{0.79}              & \textbf{0.94}                 & \textbf{0.86}                & \textbf{0.11}                                                                     & \textbf{0.79}              & \textbf{0.86}                 & \textbf{0.83}                & \textbf{0.55}                                                                     & \textbf{0.97}              & \textbf{0.80}                 & \textbf{0.88}               
\\ \bottomrule
\end{tabular}}
\vspace{-0.4cm}
\end{table*}

\Cref{tab:obj_eval_ObSLAM} compares HGG against OA-SLAM and VOOM under the ORB-SLAM2 front-end both baselines require. The results confirm the representational gap identified in \Cref{sec:ob-slam}: both model an object as a fixed parametric surface, a quadric or ellipsoid, which carries no distribution over its position or extent. Two consequences appear directly in the map. A centroid recovered by quadric fitting is weakly constrained, so position error is high, at \SIrange{0.23}{1.53}{\metre} for OA-SLAM and \SIrange{0.41}{0.76}{\metre} for VOOM against \SIrange{0.06}{0.55}{\metre} for HGG. An ellipsoid also cannot represent an object whose shape is not ellipsoidal, so recall is low, giving F1 of 0.19$\sim$0.40 and 0.08$\sim$0.33 against 0.83$\sim$0.88. Subway isolates the failure: OA-SLAM attains precision 1.00 at recall 0.10, an association conservative enough to confirm only objects observed from many viewpoints. The gap is not one of tuning or sensor quality but of the node primitive: a surface with no distribution attached can be neither compared distributionally nor weighted by its evidence.


\begin{table}[!htb]
\caption{Evaluation on the effect of NIW posterior belief in our scene graph representation.}
\label{tab:niw_ab}
\setlength{\tabcolsep}{6.5pt}
{\footnotesize
\begin{tabular}{lclllll} 
\toprule
\textbf{Sequence}                                                                   
& \multicolumn{1}{l}{\textbf{Dataset}}                                                
& \textbf{Method}   & $\Delta_p$ & \textbf{R}        & Prec.     & F1      \\ \midrule
\multirow{6}{*}{HGG}                                                    & \multirow{2}{*}{Subway} & w/o NIW  & 1.21   & \textbf{0.97}  & 0.58            & 0.73              \\
&                                                                            
& with NIW & \textbf{0.55}     & \textbf{0.97}  & \textbf{0.80} & \textbf{0.88} 
\\ \cmidrule{2-7}
& \multirow{2}{*}{\begin{tabular}[c]{@{}c@{}}In-house- \\ 02\end{tabular}}  
& w/o NIW  & 0.21           & 0.62          & 0.65          & 0.63              \\
&                                                                            
& with NIW & \textbf{0.05}  & \textbf{0.92} & \textbf{0.85} & \textbf{0.88}    
\\ \cmidrule{2-7}
& \multirow{2}{*}{MCDNTU}                                                    
& w/o NIW  & 0.87           & 0.83          & 0.69          & 0.74              \\
&                                                                            
& with NIW & \textbf{0.75}  & \textbf{0.84} & \textbf{0.82} & \textbf{0.82}          
\\ \midrule
\multirow{4}{*}{\begin{tabular}[c]{@{}l@{}}HGG+\\ GT-odom\end{tabular}} & \multirow{2}{*}{Subway} & w/o NIW  
& 0.61               & 0.94              & 0.93              & 0.94              \\ 
&                                                                            
& with NIW & \textbf{0.10}     & \textbf{0.98}     & \textbf{0.97}   & \textbf{0.97}  
\\ \cmidrule{2-7}
& \multicolumn{1}{l}{\multirow{2}{*}{MCDNTU}}                                
& w/o NIW  & 0.58           & 0.83          & 0.73          & 0.77          \\
& \multicolumn{1}{l}{}                                                       
& with NIW & \textbf{0.37}  & \textbf{0.83} & \textbf{0.87} & \textbf{0.84} 
\\ \bottomrule
\end{tabular}}
\vspace{-0.4cm}
\end{table}

\subsection{Module Evaluation}

\subsubsection{Creation}\label{sec:niw_abl}
\Cref{sec:higogo} locates the width of the node's belief in the NIW posterior, which two operations consume: the conjugate merge of \eqref{eq:merge} and the factor weight $\Omega_{\text{obj}}^{(m,i)}$ of \eqref{eq:fgo}.
This ablation strips the belief from both, replacing the merge with a symmetric average and making the factor weight uniform, so each node is reduced to the Gaussian $\mathcal{N}(\mu, \Sigma)$ with no posterior over it; the rest of the system is left in place. \Cref{tab:niw_ab} reports three settings from \Cref{tab:obj_eval}, Subway, In-house-02, and MCD-NTU, with Subway and MCD-NTU repeated under ground-truth odometry, which the in-house platform does not provide.

Position error and precision are what move: the NIW posterior belief cuts errors by 55\%, 76\%, and 14\% across the three settings and raises precision by 22, 20, and 13 points. Both follow from one mechanism. Without a belief, a single glimpse at range enters the average with the weight of an object resolved over many views, pulling the merged mean toward the poorer estimate, while the between-means term of \eqref{eq:merge} inflates the extent instead of composing two partial views. Such a node either falls outside the matching radius or fails to consolidate with the observation that should have absorbed it, and is counted as a false positive: a ghost produced by fusion rather than by association. Recall is correspondingly unchanged on Subway and MCD-NTU, since whether an object is instantiated at all is decided by the Gaussian and the distributional gate, both retained in the ablated system. 
In-house-02 is the exception, where recall falls from 0.92 to 0.62 alongside precision. Stripped of its belief, the node's covariance is an uncalibrated shape. Where the labels are predicted and most first views are partial, that corrupts not only where an object is placed but whether it is recovered at all, so the loss shows in both measures rather than in precision alone.
Ground-truth poses do not substitute for the belief: with drift removed, restoring it still cuts error from \SI{0.61}{\metre} to \SI{0.10}{\metre} on Subway and from \SI{0.58}{\metre} to \SI{0.37}{\metre} on MCD-NTU, so the belief governs how observations of one object combine rather than compensating for trajectory error. On Subway the belief matters more than the poses: HGG with VIO and the belief reaches \SI{0.55}{\metre}, against \SI{0.61}{\metre} with ground-truth poses and without it.

\subsubsection{Alignment}

\begin{table}[t]
\centering
\caption{Graph Alignment Evaluation (following~\cite{chen2026opensga}).}
\label{tab:scannetsg_s2s}
\def\arraystretch{1.15}
\setlength{\tabcolsep}{5pt}
\begin{tabular}{lcccc}
    \toprule
    \textbf{Method}
        & \textbf{Accuracy}
        & \textbf{Precision}
        & \textbf{Recall}
        & \textbf{F1-score}
         \\
    \midrule
    SG-Reg~\cite{liu2025sg}
        & 0.056 & 0.171 & 0.078 & 0.107  \\
    ROMAN~\cite{peterson2025roman}
        & 0.430 & 0.505 & 0.268 & 0.336  \\
    VLM CosSim~\cite{chen2026opensga}
        & 0.461 & 0.370 & \textbf{0.906} & 0.502  \\
    VLM+Bert CosSim~\cite{chen2026opensga}
        & 0.518 & 0.400 & \ul{0.898} & 0.530  \\
    OpenSGA-L~\cite{chen2026opensga}
        & 0.642 & 0.502 & 0.713 & \ul{0.568}  \\
    OpenSGA-H~\cite{chen2026opensga}
        & \textbf{0.681} & \textbf{0.547} & 0.684 & \textbf{0.589}  \\
    HGG 
        & \ul{0.680} & \ul{0.540} & 0.624 & \ul{0.568}  \\
    \bottomrule
\end{tabular}
\vspace{-0.5cm}
\end{table}

We evaluate whether the alignment of \Cref{sec:Alignment} returns the correct node correspondences between two graphs. The ScanNet-SG benchmark measures this directly, on independently segmented submaps with no temporal relation, so correspondence estimation runs without the odometry initialization $\Tf_\text{init}$ it receives in the incremental pipeline and without any prior on the transform. Evaluation is zero-shot on the Subscan test set, with no training on scene graph alignment data, whereas OpenSGA~\cite{chen2026opensga}, the leading method on this benchmark, trains a matching network on top of pretrained features and reports a lightweight (OpenSGA-L) and a high-performance (OpenSGA-H) variant.

\Cref{tab:scannetsg_s2s} reports the results. HGG reaches F1 0.568, matching OpenSGA-L and above every method that does not train on the benchmark, including VLM+Bert CosSim at 0.530, which carries richer semantic embeddings but no geometric structure. Its accuracy of 0.680 exceeds OpenSGA-L (0.642) and is on par with OpenSGA-H (0.681). 
The remaining F1 gap to OpenSGA-H is recall, and it falls on the inter-class correspondences that the IoU-defined ground truth includes (27.4\% of pairs). We relax the hard same-label constraint of Section IV-B1 to a soft VLM similarity cost here so that such pairs remain admissible, but the cost is designed for same-class correspondence and still ranks a same-label candidate above a cross-class one at comparable geometry.
The distributional cost of \eqref{eq:wass} therefore transfers to a task and a sensor it was not designed for, and what limits it is the deterministic semantic channel rather than the geometry. The same cost is ablated against centroid radius matching and bounding-volume IoU in the supplementary material~\cite{supplementary}, where F1 on KITTI falls from 0.91 to 0.62 and 0.71, respectively.


\begin{table}[t]
\centering
\caption{ATE Evaluation (m).}
\label{tab:ate}
\small
\setlength{\tabcolsep}{3pt}
\renewcommand{\arraystretch}{1.05}
{\footnotesize
\begin{tabularx}{\columnwidth}{l*{3}{>{\centering\arraybackslash}X}}
\toprule
\multicolumn{4}{c}{\textit{uHumans2}}\\
Method & Apartment & Office & Subway\\
\midrule
OA-SLAM~[29]      & 0.14 & 0.39 & \underline{0.45}\\
VOOM~[30]         & 0.14 & 0.42 & 0.47\\
\midrule
Kimera-RPGO~[3]   & \textbf{0.07} & 0.46 & 1.68\\
Kimera-PGMO~[3]   & \underline{0.08} & \underline{0.21} & 1.47\\
\midrule
HGG               & 0.11 & \textbf{0.17} & \textbf{0.52}\\
\bottomrule
\end{tabularx}}

\vspace{2pt}
{\footnotesize
\begin{tabularx}{\columnwidth}{l*{4}{>{\centering\arraybackslash}X}}
\toprule
\multicolumn{5}{c}{\textit{Outdoor, KISS-ICP front-end}}\\
Method & KITTI & KITTI-C & MCD-NTU & MCD-KTH   \\
\midrule
KISS-ICP~[88]   & 8.85 & 45.73 & 2.19 & 1.16\\
SG-SLAM~[23]    & \underline{3.54} & 18.36 & \underline{1.04} & 1.51\\
KISS-SLAM~[89]  & \textbf{3.06} & \underline{7.76} & 1.48 & \underline{0.67}\\
HGG             & 4.58 & \textbf{2.35} & \textbf{0.68} & \textbf{0.48}\\
\bottomrule
\end{tabularx}}
\end{table}

\subsubsection{Correction}

\paragraph{ATE Evaluation}
\Cref{tab:ate} reports trajectory error, grouped so that each comparison holds the odometry input fixed. Indoors, HGG is best on Office and Subway and within 0.04,m of the best on Apartment, improving on the Kimera pipelines by 65\% on Subway. Outdoors, HGG is best on both MCD campuses, improving on the next best system by 34\% and 28\%, and comparable to KISS-SLAM and SG-SLAM on KITTI. KITTI-Carla is the clearest case: HGG reduces the 45.73,m error of its own front-end to \SI{2.35}{\metre}. Map accuracy is therefore largely independent of front-end quality, which matters for platforms where a strong odometry pipeline is unavailable.

\paragraph{Ablation Study on Correction }\label{sec:ablation}

\begin{table}[!htb]
\centering
\caption{Effect of EM optimization and loop closure (LC)}
\label{tab:abl_opt}
{\footnotesize
\begin{tabular}{ccrrrr} 
\toprule
EM & LC & Chamfer (m) $\downarrow$ & Map Acc. (m) $\downarrow$ & F1 $\uparrow$  & ATE (m) $\downarrow$ \\ \midrule
$\checkmark$  &               & 0.72        & 0.49         & 0.63 & 2.16    \\
   & $\checkmark$             & 1.11        & 0.97         & 0.82 & 0.68    \\
$\checkmark$  & $\checkmark$  & 0.72        & 0.49         & 0.82 & 0.68    \\
\bottomrule
\end{tabular}}
\vspace{-0.3cm}
\end{table}

Correction is ablated by stage. The EM step of \Cref{sec:em} is the channel through which later observations within sliding window revise a node built from a partial view, and loop closure enforces global consistency; \Cref{tab:abl_opt} enables each in turn on MCD-NTU. The factor graph of \Cref{sec:fgo} runs in every configuration, and only the EM step and loop closure, with the pose graph optimization it feeds, are toggled. Map fidelity is measured against the survey-grade Leica RTC360 reference following the PALoc/MapEval protocol~\cite{hu2024paloc, hu2025mapeval}: the reference is cropped to points within \SI{3}{\metre} of the estimate, so the metric reflects reconstruction quality rather than traversal coverage, and the two maps are then aligned. We report the mean bidirectional Chamfer distance and the map accuracy over every node, stuff classes included, since the geometry graph is built identically for a road or a fa\c{c}ade as for a discrete object; vegetation is excluded, as foliage offers no stable surface for comparison across scan sessions.

The two stages act on disjoint quantities. EM cuts the Chamfer distance from \SI{1.11}{\metre} to \SI{0.72}{\metre} and the accuracy error from \SI{0.97}{\metre} to \SI{0.49}{\metre}, by 35\% and 49\%, and leaves ATE and F1 untouched, since it revises only $\theta^g_i$, under a fixed node count and topology, while the factor graph holds internal geometry fixed. Loop closure moves the complementary pair, ATE from \SI{2.16}{\metre} to \SI{0.68}{\metre} and F1 from 0.63 to 0.82, and leaves map fidelity unchanged. Drift is what separates them: without loop closure the trajectory diverges over a campus-scale revisit and displaces whole objects in the global frame, and a node displaced past the matching radius is charged as a miss or a false positive, whereas the geometry inside an object is refined in the object's own frame over a sliding window and measured after alignment, so it is indifferent to that displacement. This is the decoupling of \Cref{sec:Optimization} observed rather than assumed. These absolute figures also test \Cref{sec:foundation} directly: the metric map here is extracted from the graph rather than built before it, and the geometry so extracted lies within \SI{0.72}{\metre} Chamfer distance and \SI{0.49}{\metre} accuracy of a survey-grade terrestrial scan, a residual on the scale of the \SI{0.68}{\metre} trajectory error of the same configuration (\Cref{tab:abl_opt}) rather than of the object geometry.
\vspace{-0.3cm}
%
\subsection{Downstream Tasks}
\begin{table}[!htb]
\centering
\caption{Language Grounded Object Retrieval Results. HGG (centroid edges) replaces the distributional edge weight of \eqref{eq:rel_weights} with the Euclidean distance between centroids.}
\label{tab:D1}
\setlength{\tabcolsep}{9pt}
{\footnotesize
\begin{tabular}{lllll}
\toprule
Sequence                   & Method        & R@1  & R@2  & R@3   \\ \midrule
\multirow{2}{*}{Apartment} & Conceptgraphs & 0.35 & 0.41 & 0.59 \\
                           & HGG (centroid edges) & 0.47 & 0.82 & \textbf{0.94} \\
                           & HGG        & \textbf{0.65} & \textbf{0.88} & \textbf{0.94} \\ \cmidrule{2-5}
\multirow{2}{*}{Office}    & Conceptgraphs & 0.20  & 0.47 & 0.47  \\
                           & HGG (centroid edges) & \textbf{0.47} & 0.67 & \textbf{0.93} \\
                           & HGG        & \textbf{0.47} & \textbf{0.87} & \textbf{0.93}   \\ \cmidrule{2-5}
\multirow{2}{*}{Subway}    & Conceptgraphs & 0.07 & 0.29 & 0.36  \\
                           & HGG (centroid edges) & 0.21 & 0.79 & 0.86 \\ 
                           & HGG        & \textbf{0.43} & \textbf{0.93} & \textbf{0.93}   \\ 
\bottomrule
\end{tabular}}
\vspace{-0.3cm}
\end{table}
We evaluate HGG in two settings: language-grounded object retrieval tests semantic and relational reasoning, while real-world robotic navigation tests whether the same representation can directly support action without separate dense metric map.
\subsubsection{Language-Grounded Object Retrieval}
We evaluate language-grounded object retrieval on the three uHumans2 sequences against ConceptGraphs, an open-vocabulary system built for this task. Fifteen queries per sequence span unique-instance, spatial-relational (``the chair next to the computer''), and instance-disambiguation categories, each authored against the ground-truth graph so that exactly one instance satisfies it, and a retrieval counts only if that instance is returned. Both systems use the same CLIP model and checkpoint, isolating the effect of the scene-graph representation. For relational and disambiguation queries, the graph gates the candidates to $\E^o$ neighbors of the anchor, ordered by $w_{i,j}$ in \eqref{eq:rel_weights}; the full protocol is given in the supplementary material~\cite{supplementary}.

\Cref{tab:D1} reports recall at rank $k$. HGG reaches R@1 of 0.65, 0.47, and 0.43 across the three scenes against 0.35, 0.20, and 0.07 for ConceptGraphs, and by rank three returns 0.93 to 0.94 of the targets against 0.36 to 0.59. With the encoder shared, the difference reflects the underlying map: missed object instances and misplaced spatial arrangements directly limit retrieval. 
Replacing $w_{ij}$ of \eqref{eq:rel_weights} with the centroid distance, keeping the encoder fixed, leaves R@3 unchanged as $\mathcal{E}^o$ but drops R@1 to 0.47, 0.50, and 0.21 (HGG* in \Cref{tab:D1}): scoring the relation against the objects' extent rather than their centers is what orders it correctly.
Office and Subway are the harder scenes, with many same-class objects in near-identical repeated arrangements, so relational and disambiguation queries admit several candidates at once; HGG still resolves half of their queries at rank one.

\subsubsection{Real-World Language-Grounded Navigation}\label{sec:navigat}
Retrieval establishes that the graph identifies the target; we next test whether it is sufficient for action. We deploy HGG on a Boston Dynamics Spot equipped with Manifold Odin1~\cite{odin1} and issue three sequential language-grounded tasks, including two relational commands, from an indoor corridor to an outdoor bench. A task is successful when Spot stops within $1.0\,\mathrm{m}$ of the target, and all three tasks succeed. Traversable space is extracted directly from the geometry graphs without a dense metric map or retained point cloud, while the scene graph provides the semantics, distributional geometry, object extents, and evidence counts required for target identification, relational disambiguation, and goal placement. The experiment closes the loop with our representation: the scene graph does not need a geometric foundation; with the right primitive, it is one. Further experimental details and videos are provided in the supplementary material~\cite{supplementary}.


\section{Discussion}\label{sec:discussion}
We discuss two aspects left open by our instantiation of the probabilistic scene graph formulation: symbolic uncertainty and representation granularity.
\vspace{-0.25cm}
\subsection{Symbolic Uncertainty} 
The formulation in \Cref{def:psg} distinguishes uncertainty in the symbolic structure of the graph from uncertainty in the geometric parameters attached to its entities. We explicitly models the geometric belief through the Normal-Inverse-Wishart posterior, while the symbolic factor is instantiated deterministically as described in \Cref{sec:foundation}. Consequently, HGG represents uncertainty in the location and extent of an established entity, but not uncertainty in whether an entity exists, which observations should be associated with the same entity, or which semantic interpretation is correct. These uncertainties are fundamentally different from geometric uncertainty because they alter the graph structure itself rather than the parameters of an established node. A fully probabilistic scene graph would therefore need to maintain uncertainty over these symbolic variables and propagate it through creation, association, and correction, allowing competing graph hypotheses to be revised as evidence accumulates. Thus, the current system should be viewed as instantiating the probabilistic geometric component of the broader formulation while leaving symbolic uncertainty as an orthogonal extension in our future work.

\vspace{-0.2cm}
\subsection{Hierarchy and Granularity} %
The probabilistic scene graph formulation also does not prescribe the spatial granularity and hierarchy at which entities should be represented. This is important because granularity determines both the computational scope of graph inference and the type of information that an entity can express: coarse entities provide compact and stable support for global reasoning~\cite{gu2024conceptgraphs}, whereas fine entities preserve greater geometric detail at increased state and association complexity. The two levels in our work represent the same entity at different resolutions rather than independent maps: the object graph provides the abstraction required for efficient global reasoning, while the geometry graph supplies the spatial detail required for local reconstruction and refinement. This hierarchy is a design choice rather than a restriction of the underlying formulation, and illustrates how a probabilistic scene graph can decouple the granularity required for global semantic reasoning from that required for geometric reconstruction.
The same construction extends upward as easily as downward, grouping objects into regions rather than dividing them into parts, as sketched in \Cref{sec:Initialization}. What a further level would require is not more geometry, but a belief over symbolic factor which provide information on entities belong together.
More adaptive choices of granularity and hierarchy could further extend this framework, but would require jointly reasoning about the representation scale and the graph structure itself.

\section{Conclusion} 
\label{sec:conclusion}

We introduced the probabilistic scene graph, a representation whose output is a posterior over graphs rather than a single graph, and what creation, alignment, and correction must preserve for a node to remain a belief. We instantiated the PSG, whose object nodes are full-covariance Gaussians on $\spd$ carrying an evidence-calibrated Normal-Inverse-Wishart posterior, refined by a geometry graph and joined by distributional-distance edges. On it we built a purely graph-based coarse-to-fine alignment needing no dense map, and a nested EM--factor-graph optimization splitting global arrangement from local geometry as the Fisher information structure prescribes. Carrying its own geometry, the graph replaces the dense metric map rather than abstracting from it: across six datasets, Our system runs at sensor rate with near-constant memory at state-of-the-art object accuracy and zero-shot alignment. Extending the same requirements to the semantic, existence, and correspondence factors is what the definition already calls for, and dynamic objects emerge from it as those whose posterior never concentrates.

\bibliographystyle{IEEEtran}
\bibliography{references}

\end{document}